\documentclass[nohyperref]{article}

\usepackage{microtype}
\usepackage{graphicx}
\usepackage{subfigure}
\usepackage{booktabs} %
\usepackage{tipa}

\usepackage{hyperref}

\usepackage[accepted]{icml2023}

\usepackage{amsmath}
\usepackage{amssymb}
\usepackage{mathtools}
\usepackage{amsthm}

\usepackage{amsmath,amsfonts,bm,bbm,siunitx}
\usepackage[utf8]{inputenc}
\usepackage{enumitem}
\usepackage{mleftright}

\def\Figref#1{Figure~\ref{#1}}

\def\eqref#1{equation~\ref{(#1)}}
\def\Eqref#1{Equation~(\ref{#1})}

\def\Algref#1{Algorithm~\ref{#1}}

\def\ceil#1{\lceil #1 \rceil}

\def\1{\mathbbm{1}}

\def\va{{\bm{a}}}
\def\vb{{\bm{b}}}

\def\vg{{\bm{g}}}
\def\vh{{\bm{h}}}

\def\vm{{\bm{m}}}

\def\vx{{\bm{x}}}

\def\vz{{\bm{z}}}

\def\mA{{\bm{A}}}

\def\mD{{\bm{D}}}

\def\mF{{\bm{F}}}

\def\mH{{\bm{H}}}

\def\mM{{\bm{M}}}

\def\mP{{\bm{P}}}

\def\mS{{\bm{S}}}

\def\mW{{\bm{W}}}

\DeclareMathAlphabet{\mathsfit}{\encodingdefault}{\sfdefault}{m}{sl}
\SetMathAlphabet{\mathsfit}{bold}{\encodingdefault}{\sfdefault}{bx}{n}

\def\sA{{\mathbb{A}}}
\def\sB{{\mathbb{B}}}

\def\sP{{\mathbb{P}}}

\def\sT{{\mathbb{T}}}

\newcommand{\E}{\mathbb{E}}

\newcommand{\R}{\mathbb{R}}

\DeclareMathOperator*{\argmax}{arg\,max}

\DeclareMathOperator{\sign}{sign}

\newcommand{\our}{Globe}
\newcommand{\ourlong}{Graph-learned Orbital Embeddings}
\newcommand{\ourwf}{Moon}
\newcommand{\ourwflong}{Molecular Orbital Network}

\def\nvec{{\mathbf{R}}} %
\def\n{{\bm{R}}} %

\def\o{{\bm{L}}} %

\def\evec{{\mathbf{r}}} %
\def\e{\bm{r}} %

\def\filter{{\varGamma}}
\def\act{{\sigma}}
\def\spin{{\alpha}}

\DeclareSIUnit\bohr{\ensuremath{a_0}}
\DeclareSIUnit\hartree{\text{\ensuremath{E_\textup{h}}}}
\DeclareSIUnit\calorine{cal}
\DeclareSIUnit\kcal{\kilo\calorine}

\DeclareUnicodeCharacter{02C8}{\textprimstress}
\DeclareUnicodeCharacter{026A}{\textsci}
\DeclareUnicodeCharacter{0292}{\textyogh}
\DeclareUnicodeCharacter{0254}{\textopeno}
\DeclareUnicodeCharacter{02A4}{\textdyoghlig}

\usepackage[capitalize,noabbrev]{cleveref}

\theoremstyle{plain}

\theoremstyle{definition}

\theoremstyle{remark}

\usepackage[textsize=tiny]{todonotes}

\icmltitlerunning{Generalizing Neural Wave Functions}

\begin{document}

\twocolumn[
\icmltitle{Generalizing Neural Wave Functions}

\icmlsetsymbol{equal}{*}

\begin{icmlauthorlist}
\icmlauthor{Nicholas Gao}{tum}
\icmlauthor{Stephan Günnemann}{tum}
\end{icmlauthorlist}

\icmlaffiliation{tum}{Department of Computer Science \& Munich Data Science Institute, Technical University of Munich, Germany}

\icmlcorrespondingauthor{Nicholas gao}{\href{mailto:n.gao@tum.de}{n.gao@tum.de}}

\icmlkeywords{Graph Neural Networks, GNN, Molecules, Computational Physics, Computational Chemistry, Quantum Chemistry, Self-Generative Learning, Machine Learning for Science, Quantum Monte Carlo, Variational Monte Carlo, Neural Quantum States}

\vskip 0.3in
]

\printAffiliationsAndNotice{}  %

\begin{abstract}
   Recent neural network-based wave functions have achieved state-of-the-art accuracies in modeling \emph{ab-initio} ground-state potential energy surface. However, these networks can only solve different spatial arrangements of the same set of atoms.
   To overcome this limitation, we present \ourlong{} (\our{}), a neural network-based reparametrization method that can adapt neural wave functions to different molecules.
   \our{} learns representations of local electronic structures that generalize across molecules via spatial message passing by connecting molecular orbitals to covalent bonds.
   Further, we propose a size-consistent wave function Ansatz, the \ourwflong{} (\ourwf{}), tailored to jointly solve Schrödinger equations of different molecules. 
   In our experiments, we find \ourwf{} converging in 4.5 times fewer steps to similar accuracy as previous methods or to lower energies given the same time.
   Further, our analysis shows that \ourwf{}'s energy estimate scales additively with increased system sizes, unlike previous work where we observe divergence.
   In both computational chemistry and machine learning, we are the first to demonstrate that a single wave function can solve the Schrödinger equation of molecules with different atoms jointly.
\end{abstract}

\section{Introduction}\label{sec:introduction}
\emph{In silico} design of molecules requires accessing their quantum mechanical properties.
This requires solving the associated Schrödinger equation.
However, exact solutions are often intractable and approximations can become computationally expensive for larger and more complex systems.
In recent years, neural network-based wave functions have emerged as a promising alternative, providing accurate and well-scaling approximation solutions in $O(N^4)$ with the number of electrons $N$~\citep{hermannAbinitioQuantumChemistry2022}.
Despite their theoretical scaling, this time complexity comes with a large prefactor leading to exploding computational requirements if one screens many different molecules.
To address this limitation, \citet{gaoAbInitioPotentialEnergy2022} proposed the Potential Energy Surface Network (PESNet), a neural network wave function that generalizes across different structures.
While reducing computational costs, PESNet is limited to different spatial arrangements of the same set of atoms.

A key challenge in the generalization to arbitrary molecules is the variable number of molecular orbitals.
To resolve this, we introduce \ourlong{} (\our), a generalization of PESNet that can solve arbitrary Schrödinger equations jointly.
Like previous work, \our{} uses a two-level approach where one network represents the electronic wave function and the other reparametrizes the wave function depending on the molecule.
We resolve the issue of the dynamical numbers of molecular orbitals by embedding orbitals in 3D space.
This enables us to learn local electronic structures via spatial message passing in graph neural networks (GNNs).
For the wave function, we present the \ourwflong{} (\ourwf), the first size-consistent neural wave function.
We accomplish size consistency in two key steps, firstly, by using spatial message passing \ourwf{} focuses on local interactions, and, secondly,  by using the nuclei as anchor points for message passing.
While the first step is strictly required for size consistency, the latter enables efficient reparametrization via \our{}.

In our experiments, we find \ourwf{} accelerating convergence in joint training by up to 4.5 times and performing similarly to the attention-based PsiFormer on larger systems~\citep{vonglehnSelfAttentionAnsatzAbinitio2023}.
Further, we observe that transfers of neural wave functions to larger structures do not require additional self-consistent field (SCF) calculations.
In summary, our main contributions are:\footnote{Source code:  \href{https://www.cs.cit.tum.de/daml/globe/}{https://www.cs.cit.tum.de/daml/globe/}}
\setlist{nolistsep}
\begin{itemize}[noitemsep]
    \item \textbf{\our}, a reparametrization method for adapting neural wave functions to arbitrary molecules based on localized molecular orbital embeddings.
    \item \textbf{\ourwf}, a size-consistent neural wave function enabling generalization to larger structures, faster convergence, and accurate energies.
\end{itemize}

\section{Background}\label{sec:background}

\textbf{Notation.} %
We use the term molecule for a point cloud in $\R^3$ with charges assigned to each node.
The term `geometry' refers to different spatial arrangements associated with the same set of charges.
We use $N$ to denote the number of electrons and $M$ for the number of nuclei.
$\evec\in\R^{N\times 3}$ denotes a complete electron configuration whereas $\e\in\R^3$ denotes a single electron's position.
For nuclei, we use $\nvec\in\R^{M\times 3}$ and $\n\in\R^3$, respectively.
$Z_m$ denotes the charge of the $m$th nucleus.
We use $[\phantom{a}]$ for the concatenation of vectors, $\circ$ for the Hadamard product, $\Vert x\Vert$ for the $L_2$-norm, bold capital letters $\mA$ for matrices, bold lower case letters $\va$ for vectors, and normal face letters $a$ for scalars. 
Bracketed superscripts$^{(l)}$ index sequences, e.g., layers in a neural network.

\subsection{Quantum chemistry}
At the heart of quantum chemistry is the Schrödinger equation. 
Its time-independent form is
\begin{align}
    \mH \psi = E \psi \label{eq:schroedinger}
\end{align}
where $\psi: \R^{N\times 3}\rightarrow\R$ is the electronic wave function, $E$ the energy and the Hamiltonian operator is
\begin{align}
    \mH =& -\frac{1}{2}\sum_{n=1}^{3N}\nabla_n^2 + V(\evec),\\
    \begin{split}
        V(\evec) =& \sum_{n>m=1}^N\frac{1}{\Vert r_n - r_m\Vert}
             - \sum_{n=1}^N\sum_{m=1}^M \frac{Z_m}{\Vert r_n - R_m\Vert} \\
             &+\sum_{m>n=1}^M \frac{Z_mZ_n}{\Vert R_m - R_n\Vert},
    \end{split}
\end{align}
within the Born-Oppenheimer approximation, i.e., we approximate nuclei as particles with fixed positions.
In linear algebra, \Eqref{eq:schroedinger} is an eigenvalue problem where one wants to find the eigenfunction $\psi_0$ associated with the lowest eigenvalue $E_0$.
These are commonly called the ground-state wave function and energy, respectively.

The electronic wave function $\psi$ describes the behavior of electrons.
Note that electrons are not only specified by their spatial location $\e\in\R^3$ but also by their spin $\alpha\in\{\uparrow,\downarrow\}$.
Though, since the spins do not occur in the Hamiltonian, they can be fixed a priori~\citep{foulkesQuantumMonteCarlo2001}.
For a function to be a valid wave function, it must satisfy two criteria.
First, $\psi$ must obey the Fermi-Dirac statistics, i.e., it must be antisymmetric $\psi(\evec)=\sign(\pi)\psi(\pi(\evec))$ w.r.t. permutations of same-spin electrons $\pi$.
Second, the integral of its square must be one $\int \psi(\evec)^2 d\evec=1$.

The challenge in computational chemistry is accurately approximating the ground-state energy.
For instance, the total energy of a system can be decomposed into a mean-field energy and correlation energy, where the mean-field energy accounts for $\approx 99.5\%$ of the total energy.
To reach chemical accuracy (typically defined as \SI{1}{\kcal\per\mole}), one has to accurately estimate $> 99\%$ of the correlation energy, $>99.999\%$ of the total energy.

Most commonly, the wave function is represented by a determinant of molecular orbital functions~\citep{slaterTheoryComplexSpectra1929}:
\begin{align}
    \psi(\evec) = \det\Phi,&& \Phi_{ij} = \phi_j(r_i)
    \label{eq:hf}
\end{align}
where the determinant ensures the antisymmetry w.r.t. permutations.
The Hartree-Fock (HF) method provides a simple mean-field approximate solution to the Schrödinger equation where the molecular orbital functions $\phi^\text{HF}_i:\R^{3}\rightarrow\R$ are constructed with Linear Combinations of Atomic Orbitals (LCAO) $\varphi_j:\R^{3}\rightarrow\R$~\citep{lennard-jonesElectronicStructureDiatomic1929}, $\phi^\text{HF}_i(x)=\sum_{m=1}^{M}\sum_{n=1}^{O_m} \omega_{i,m, n}\varphi_{m,n}(x)$ with $O_m$ being the number of atomic orbitals and $\varphi_{m,n}$ being the $n$th atomic orbital function of the $m$th atom, respectively.
In matrix notation, \Eqref{eq:hf} can be written as 
\begin{align}
    \psi(\evec) = \det\Phi &= \det (\varPhi\Omega^T)& \varPhi,\Omega &\in \R^{N\times \eta}\label{eq:hf_mat}
\end{align}
with $\eta=\sum_{m=1}^{M}O_m$, $\varPhi$ being a matrix of all atomic orbital functions evaluated at every electron position, and $\Omega$ being an optimized weight matrix.

\subsection{Variational Monte Carlo}
In Variational Monte Carlo (VMC), one approximates a solution to \Eqref{eq:schroedinger} by picking a trial wave function $\psi_\theta$ parametrized by $\theta$ and iteratively minimizing the energy via gradient descent on $\theta$.
Since the eigenfunctions of $\mH$ are a complete basis, this variational optimization is an upper bound to the true ground-state energy.
By reformulating \Eqref{eq:schroedinger}, one gets
\begin{align}
    E &= %
    \frac{
        \int \psi_\theta(\evec) \mH \psi_\theta(\evec) d\evec
    }{
        \int \psi^2_\theta(\evec) d\evec
    }\\
    &= \E_{\evec\sim \psi^2_\theta}\left[\psi_\theta(\evec)^{-1}\mH\psi_\theta(\evec)\right] = \E_{\evec\sim \psi^2_\theta}\left[E_\theta(\evec)\right].
\end{align}
In contrast to \Eqref{eq:schroedinger}, we here assumed an unnormalized wave function $\psi_\theta$, thus, the normalization factor.
Further, we reformulate the integral in the second line using importance sampling.
$E_\theta$ is the so-called local energy:
\begin{align}
    E_\theta(\evec) =& \psi_\theta(\evec)^{-1}\mH\psi_\theta(\evec)\\
    \begin{split}
        =&-\frac{1}{2}\sum_{i=1}^{3N}\left[
            \frac{\partial^2 \log\vert\psi_\theta(\evec)\vert}{\partial \evec_{i}^2} +
            \frac{\partial\log\vert\psi_\theta(\evec)\vert}{\partial \evec_{i}}^2
        \right]\\
        &+ V(\evec).\label{eq:energy}
    \end{split}
\end{align}
Finally, one optimize $\psi_\theta$ via gradient descent with 
\begin{align}
    \nabla_\theta E &= \E_{\evec\sim \psi_\theta^2}\left[
        \left[
            E_\theta(\evec) - \E_{\evec\sim \psi_\theta^2}\left[E_\theta(\evec)\right]
        \right]
        \nabla_\theta \log\psi_\theta(\evec)
    \right] \label{eq:gradient}
\end{align}
where we estimate all expectations with Monte Carlo estimates using Metropolis-Hastings~\citep{ceperleyMonteCarloSimulation1977}.

Due to the few constraints imposed on wave functions, recent works used neural networks to model them~\citep{pfauInitioSolutionManyelectron2020,hermannDeepneuralnetworkSolutionElectronic2020}.
In the neural network setting, learnable many-electron orbital functions $\phi_i:\R^{3}\times\R^{N\times 3}\rightarrow\R$ implemented by permutation equivariant neural networks replace the single-electron molecular orbital functions $\phi_i^\text{HF}$ in \Eqref{eq:hf}.

\section{Related Work}\label{sec:related_work}
Traditional methods for modeling electronic wave functions have relied on Linear Combinations of Atomic Orbitals (LCAO)~\citep{lennard-jonesElectronicStructureDiatomic1929} arranged in a Slater determinant~\citep{slaterTheoryComplexSpectra1929}.
However, they cannot capture electron-electron interactions beyond a mean-field approximation. 
To address this issue, backflow transformations~\citep{feynmanEnergySpectrumExcitations1956} and Jastrow factors~\citep{jastrowManybodyProblemStrong1955} have been introduced.
Later, \citet{carleoSolvingQuantumManyBody2017} were the first to demonstrate the use of neural networks to model to quantum systems, though only for discrete spin systems.
This approach has since been improved upon by using deep neural networks for real-space electronic systems~\citep{hanSolvingManyelectronSchrodinger2019,pfauInitioSolutionManyelectron2020,hermannDeepneuralnetworkSolutionElectronic2020}.
In subsequent works, such neural networks have further refined~\citep{gerardGoldstandardSolutionsSchrodinger2022,vonglehnSelfAttentionAnsatzAbinitio2023} and adopted to different settings like pseudopotentials~\citep{liFermionicNeuralNetwork2022}, periodic systems~\citep{wilsonWaveFunctionAnsatz2022,liInitioCalculationReal2022,cassellaDiscoveringQuantumPhase2023} or diffusion Monte Carlo (DMC)~\citep{wilsonSimulationsStateoftheartFermionic2021,renGroundStateMolecules2022}.

Despite their high accuracy, neural network-based wave function models are still inherently expensive for multiple systems. Two recent concurrent approaches addressed this challenge: DeepErwin, a weight-sharing method across geometries~\citep{scherbelaSolvingElectronicSchrodinger2022}, and PESNet~\citep{gaoAbInitioPotentialEnergy2022,gaoSamplingfreeInferenceAbInitio2023}, a two-network approach that allows for joint training of several geometries, eliminating the need for retraining.
But, while the former needs retraining for each structure, the latter is limited to different spatial arrangements of the same set of atoms.

\section{Generalizing Neural Wave Functions}\label{sec:method}
\begin{figure*}
    \includegraphics[width=\linewidth]{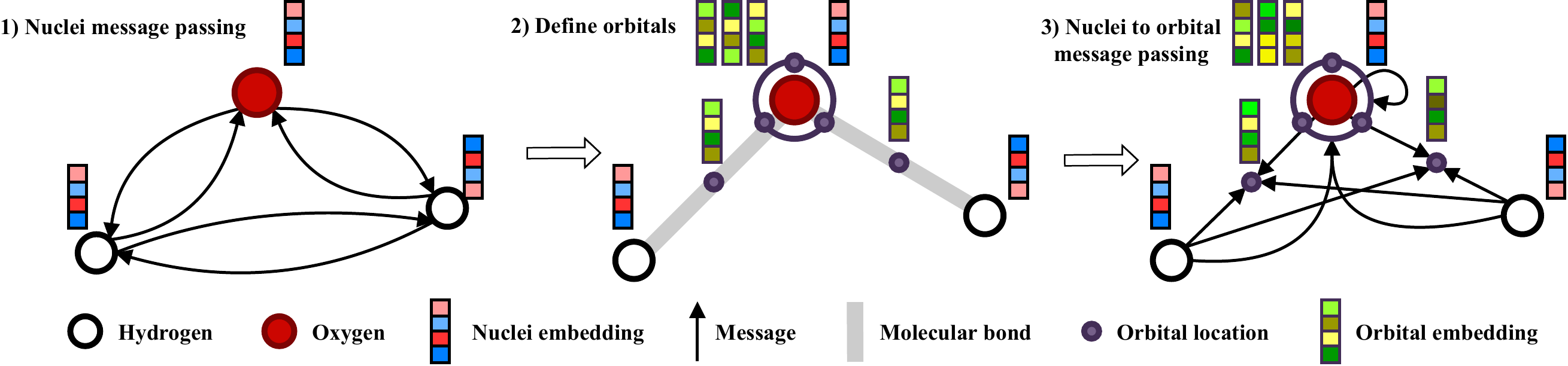}
    \vspace{-0.25in}
    \caption{
        \our{} steps.
        1) Atom embeddings are obtained by message passing between atoms. 2) Orbital locations and embeddings are determined by core orbitals and molecular bonds. 3) Orbital embeddings are updated via a unidirectional message passing. For clarity, we omitted messages from each atom to each of the three core orbitals.
    }
    \vspace{-0.15in}
    \label{fig:orb_emb}
\end{figure*}
Compared to different geometries, generalization across different molecules comes with additional difficulties as the number of atoms, electrons, and orbitals change.
To address these challenges, we identify two key desiderata that such a system should fulfill:
\setlist{nolistsep}
\begin{enumerate}[noitemsep]
    \item \emph{Invariance:} A molecule's energy is invariant to Euclidean transformations and permutations.
    Thus, a generalizing wave function should result in invariant energy estimates.
    To achieve this, the wave function must be equivariant to Euclidean transformations and nuclei permutation~\citep{gaoAbInitioPotentialEnergy2022}.
    \item \emph{Size consistency:} As most quantum mechanical interactions happen within a short distance, a molecule's energy is an extensive quantity and scales additively with its size.
    For wave functions, this implies that the wave function decomposes into a product of the individual wave functions for distant molecules.
\end{enumerate}

To incorporate Euclidean symmetries, we follow \citet{gaoAbInitioPotentialEnergy2022} by defining a PCA-based equivariant coordinate frame.
Thus, in the following all references to the electron $\evec$ and nuclei $\nvec$ positions are measured in the equivariant frame.
To accomplish size consistency, the wave function must decompose into a product of two wave functions if two systems are sufficiently separated.
In Appendix~\ref{app:size_consistency}, we show that decaying the value of molecular orbital functions $\phi_i$ to 0 far from the involved atoms is sufficient to implement this.
We achieve this by relying on local interactions between pairs of particles (electrons/atoms) that exponentially decay with distance.

Like previous work, we adopt a two-network approach.
While \ourwf{} represents the electronic wave function $\psi_\theta$, \our{} acts solely on the nuclei and adapts the former's parameters to the molecule.

\subsection{\ourlong{} (\our)}\label{sec:reparam}
\our{}'s task is to reparametrize the wave function depending on the molecular structure, i.e., it only acts on the atoms and does not consider electrons.
To perform such a reparametrization, it must extract local electronic structure information from the atomic point cloud.
Further, we must parametrize $N$ molecular orbital functions $\phi_i$, see \Eqref{eq:hf}, which poses a challenge as their number depends on the number of electrons rather than atoms.
As illustrated in Figure~\ref{fig:orb_emb}, we achieve both in a three-step procedure.
First, we learn about atomic neighborhoods via message passing.
Next, we localize orbitals and, finally, learn orbital embeddings via unidirectional message passing.

\textbf{Message-passing network.} Our message-passing network relies on the use of continuous filter convolutions~\cite{schuttSchNetDeepLearning2018}.
We initialize the node embeddings by a charge embedding $\vh^{\text{atom}(0)}_i=\mF^\text{atom}_{Z_i}$ and iteratively update them through message-passing as
\begin{align}
    \vh^{(l+1)}_i &= f^{(l)}(\vh^{(l)}_i, \vm^{(l)}_i),\\
    \vm^{(l)}_i &= \frac{1}{\nu_{\n_i}^\nvec} \sum^{M}_{j=1} g^{(l)}(\vh^{(l)}_i, \vh^{(l)}_j) \circ \filter^{(l)}(\n_i - \n_j), \\
    \nu_x^\mathcal{N} &= 1 + \sum_{y\in\mathcal{N}} \exp\left(-\frac{\Vert x-y\Vert^2}{\sigma^2_\text{norm}}\right)\label{eq:norm}
\end{align}
where $f^{(l)}$ and $g^{(l)}$ are implemented by MLPs, $\filter$ are spatial filters and $\nu$ is a spatial normalization with $\sigma_\text{norm}$ being a learnable parameter.
By multiplying elementwise with spatial filters rather than concatenating with them, as done in \citet{gaoAbInitioPotentialEnergy2022}, we decay long-range interactions between atoms and strengthen local interactions.
Further, instead of averaging over all atoms, we normalize the message by a learnable normalization factor to account for the size of its neighborhood.

\textbf{Spatial filters.}
To model arbitrary wave functions, we must break euclidean symmetries in our reparametrization~\citep{gaoAbInitioPotentialEnergy2022}.
Previous work used positional encodings relative to the center of mass to achieve this.
But, as the center of mass is an inherently global property, we instead break the symmetries in our spatial filters enforcing locality.
Instead of being radial, our filters operate on the full three-dimensional space by constructing them as a Hadamard product of a Gaussian envelope and an MLP on the three-dimensional input:
\begin{align}
    \filter^{(l)}(\vx) =& \mW^{(l)}\beta(\vx), \label{eq:filter}\\
    \begin{split}
        \beta(\vx) =& \mW^\text{env} \left[ \exp\left(-\left(\frac{\Vert \vx\Vert}{\varsigma_i}\right)^2\right) \right]_{i=1}^{D}\\
        &\circ \left(\act\left(\vx\mW^{(1)}+\vb^{(1)}\right)\mW^{(2)} + \vb^{(2)}\right)
    \end{split}
\end{align}
with $D$ being the number of envelope ranges $\varsigma_i$, and $\act$ being an activation function. 
While a combination of spherical harmonics and radial basis functions achieves similar symmetry breaking~\citep{gasteigerGemNetUniversalDirectional2021,zitnickSphericalChannelsModeling2022}, we found such freely learnable filters to perform better.

\textbf{Orbital localization.}
A key challenge in adapting a wave function to arbitrary molecules is the molecular orbital functions $\phi_i$ as their number is not fully specified by the number of atoms but by the number of electrons.
While generating one molecular orbital function per electron seems like an intuitive solution, this would cause two rows \emph{and} columns in \Eqref{eq:hf} to permute if two electrons permute, resulting in a permutation symmetric rather than an antisymmetric function.
Thus, the orbital functions must be independent of the actual electrons.
One could generate the orbitals by a global graph embedding, e.g., via an RNN, but such a construction does not preserve locality and behaves unpredictably to changes in the nuclei.

We avoid such global constructions, by assigning each molecular orbital a location $\o_i\in\R^3$ and learning the parameters of the associated orbital function $\phi_i$ via message passing.
To localize the orbitals, we distinguish between core and valence orbitals as core orbitals tend to interact little with other atoms~\citep{foulkesQuantumMonteCarlo2001}.
For each atom type, we define its valency by the number of bonds it can form, e.g., for hydrogen one, for carbon four, for oxygen two, etc. 
The number of core orbitals for the $i$th atom is then $\frac{Z_i-V_i}{2}$ with $V_i$ being the valency of the $i$th atom.
These core orbitals are located at the same location as the nuclei.
To determine the valence orbitals, we identify covalent bonds and locate the orbital in the center of that bond.
We do this by iteratively picking the pairs of atoms closest to each other where each atom has at least one unpaired electron left.
An example of our localized orbitals is depicted in Figure~\ref{fig:orb_emb}.
In Appendix~\ref{app:algorithm}, we provide a full definition of the algorithm.

While obtaining the orbital locations $\o_i$, we also define their types $T_i$.
Where the order and the charge of the nucleus define the core orbital types, the bond's cardinality defines the valence orbital types.
This categorial distinction avoids identical embeddings for two orbitals at the same location.

\textbf{Orbital embedding.} We initialize the orbital embeddings via their types $\vh^{\text{o} (0)}_i=\mF^\text{o}_{T_i}$ and iteratively update them with an architecturally identical GNN as used for the atoms but with unidirectional message passing from the atoms to the orbitals.
Here, we avoid bidirectional message passing as the orbital structure is wholly inferred from the nuclei and, thus, carries no additional geometric information.

\textbf{Parameter estimation.}
Changes in a molecule's structure manifest in wave function parameters that depend either on atoms, orbitals, or a combination of both.
Thus, \our{} updates these parameters via their respective embeddings ($\vh^\text{a}, \vh^\text{o}, \vh^\text{a-o}$).
Before generating parameters via individual MLPs, we pass them through shared MLPs and LayerNorms~\citep{baLayerNormalization2016}.
To ensure that the wave function converges to a product for distant systems (Desiderata 2), we define atom-orbital embeddings as
\begin{align}
    \vh^\text{a-o}_{i,m} = \left[\vh^{\text{a} (L)}_m, \vh^{\text{o} (L)}_i\right]\mW \circ \filter^{\text{a-o}}(\n_m-\o_i).
\end{align}
where the spatial filters vanish the contribution of the $m$th atom to the $i$th orbital with increasing distances.

\subsection{\ourwflong{} (\ourwf)}\label{sec:our_wf}
\begin{figure*}
    \includegraphics[width=\linewidth]{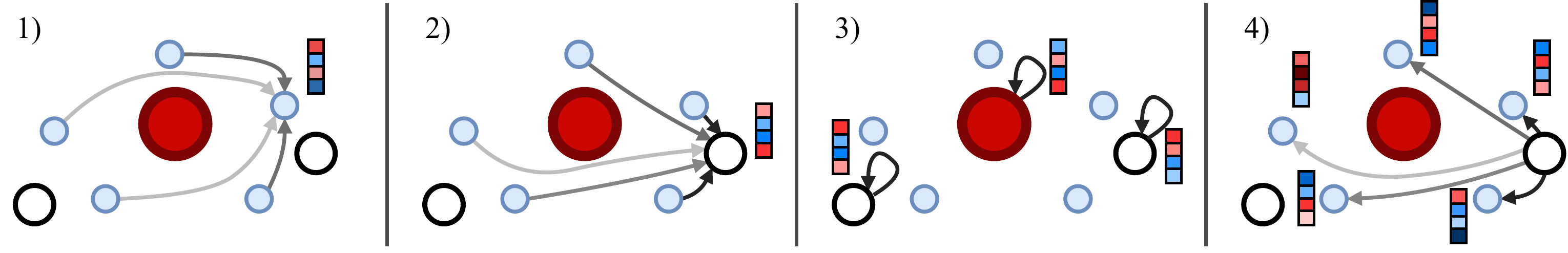}
    \vspace{-0.25in}
    \caption{
        Illustration of \ourwf.
        1) We initialize electron embeddings by aggregating their local neighborhood of electrons.
        2) Nuclei aggregate electron embeddings via message passing.
        3) Nuclei embeddings are iteratively updated.
        4) Nuclei embeddings are structurally diffused towards the electrons via message passing.
        For clarity, we omitted most messages in 1), 2), and 4).
    }
    \vspace{-0.15in}
    \label{fig:wf}
\end{figure*}
\ourwf{} represents the electronic wave function $\psi_\theta$, but, unlike previous work, \ourwf{} encourages local interactions via spatial message passing and avoids strong global interactions~\citep{pfauInitioSolutionManyelectron2020,vonglehnSelfAttentionAnsatzAbinitio2023}.
While PauliNet already represents a GNN-based wave function, it relies heavily on HF calculations and does not reach similar accuracy~\citep{hermannDeepneuralnetworkSolutionElectronic2020,gerardGoldstandardSolutionsSchrodinger2022}.

To avoid many expensive message-passing steps between electrons and nuclei, we simplify the message-passing structure.
\Figref{fig:wf} provides a conceptual overview of \ourwf.
First, we encode the local electronic neighborhood for each electron.
Next, nuclei aggregate electronic structure information.
The nuclei embeddings are then iteratively updated and, lastly, diffused to the electrons.
The basic functional form of \ourwf{} follows a Slater-Jastrow wave function
\begin{align}
    \psi_\theta(\evec) &= \exp(J(\evec))\sum_{k=1}^{K}w_k\det \Phi^k,
\end{align}
i.e., a product of a permutation invariant Jastrow factor and a weighted sum of Slater determinants.
As Jastrow factor, we additively combine the Jastrow factors from \citet{gaoSamplingfreeInferenceAbInitio2023} and \citet{vonglehnSelfAttentionAnsatzAbinitio2023}.
In the following, we use bars $\bar{a}$ for atom parameters, tildes $\tilde{a}$ for orbital parameters, and both $\tilde{\bar{a}}$ for atom-orbital interaction parameters.
These are the parameters that are updated through \our{}.
For clarity, we omit the Jastrow factor, residual connections, and normalization coefficient definitions here and refer the reader to Appendix~\ref{app:wave_function} for detailed descriptions.

\textbf{Embedding.} As initial features, we use the pairwise distances between electrons and nuclei $\vg^\text{e-n}_{ij}=[\e_i-\n_j, \Vert \e_i -\n_j\Vert]$, and electron and electrons $\vg^\text{e-e}_{ij}=[\e_i-\e_j, \Vert \e_i - \n_j\Vert]$.
As illustrated in Figure~\ref{fig:wf}, we initialize the electron embeddings by a single electron-electron message-passing step
\begin{align}
    \vh^{\text{e} (0)}_i &= 
        \sum_{j=1}^{N} \act\left(\vg^\text{e-e}_{ij}\mW^{\delta_{\spin_i}^{\spin_j}}\right)
        \circ \filter^{\delta_{\spin_i}^{\spin_j}}(\Vert \e_i-\e_j) \Vert
    \mW\label{eq:elec_init}
\end{align}
where functions and matrices superscripted by the Kronecker delta $\delta_{\spin_i}^{\spin_j}$ indicate different weights.
Here, we again use the spatial filters to decay interactions from far-apart particles.
Like \citet{gaoAbInitioPotentialEnergy2022}, we construct electron-nuclei interaction embeddings by combining electron embeddings $\vh^{\text{e} (0)}_i$, nuclei embeddings $\bar{\vz}_m$, and their distance $\vg^\text{e-n}_{im}$ via
\begin{align}
    \vh^{\text{e-n} (0)}_{im} &= \act\left(\vh^{\text{e} (0)}_i  + \bar{\vz}_m + \vg^\text{e-n}_{im}\bar{\mW}_m\right).
    \label{eq:elec_nuc_pair}
\end{align}
As the second step in Figure~\ref{fig:wf}, these embeddings are then aggregated towards electrons and nuclei via spatial message passing while keeping separate embeddings for each spin state $\spin\in\{\uparrow,\downarrow\}$ per nuclei:
\begin{align}
    \vh^{\text{n}\spin (1)}_m &= 
    \sum_{i\in \sA^\spin} \vh^{\text{e-n}}_{i,m} \circ \bar{\filter}_m^\text{n}(\e_i-\n_m),\label{eq:nuc_emb} \\
    \vh^{\text{e} (1)}_i &= 
    \sum_{m=1}^{M} \vh^{\text{e-n}}_{i,m} \circ \bar{\filter}_m^\text{e}(\e_i-\n_m)\label{eq:elec_emb}
\end{align}
where $\sA^\spin$ is the index set of electrons with spin $\spin$ and $\bar{\filter}$ being the spatial filters from \Eqref{eq:filter} with atom-parameters, see Appendix~\ref{app:wave_function}.
By using message passing instead of concatenation as commonly done in single-molecule works~\citep{vonglehnSelfAttentionAnsatzAbinitio2023}, we achieve invariance to nuclei permutations, see Desiderata 1 in Section~\ref{sec:method}.

\textbf{Update.} We iteratively update the nuclei embeddings
\begin{align}
    \vh^{\text{n}\spin (l+1)}_m &= \vh^{\text{n}\spin (l)}_m + \act([\vh^{\text{n}\spin (l)}_m, \vh^{\text{n}\hat{\spin} (l)}_m]\mW^{(l)} + \vb^{(l)})
\end{align}
where $\hat{\spin}$ denotes the opposing spin of $\spin$.
For efficiency reasons, we do not perform message passing between nuclei here as we found it to have no significant impact.

\textbf{Diffusion.} After $L$-many update steps, a single message-passing step diffuses the nuclei embeddings to the electrons
\begin{align}
    \vh^{\text{e} (L)}_i &= \act(\vh^{\text{e} (0)}_i\mW + \vm_i),\\
    \begin{split}
        \vm_i &=
        \sum_{m=1}^{M}\left(\left[\vh^{\text{n}\spin_i (L)}_m, \vh^{\text{n}\hat{\spin}_i (L)}_m\right]\mW+ \vb\right) \\
        &\phantom{=}\circ \bar{\filter}_m^\text{diff}(\e_i-\n_m)\label{eq:diff}
    \end{split}
\end{align}
with $\spin_i$ denoting the spin of the $i$th electron.
The spatial filters in this step enable the network to learn different directional messages which are important in modeling directional wave functions, e.g., the excited states of the hydrogen atom.

\textbf{Orbital construction.} After diffusion, we construct restricted orbitals like \citet{gaoSamplingfreeInferenceAbInitio2023} with adaptive orbital and envelope parameters:
\begin{align}
    \begin{split}
        \phi^{k}_{i}(\e_j) =& \left((\tilde{w}_i^{k\delta_{\spin_i}^{\spin_j}})^T\vh^{\text{e} (L)}_j + \tilde{b}_j^{k\delta_{\spin_i}^{\spin_j}} \right)\\
        &\sum_{m=1}^{M}\tilde{\bar{\pi}}_{im}^{k\delta_{\spin_i}^{\spin_j}} \exp(-\tilde{\bar{\sigma}}^{k\delta_{\spin_i}^{\spin_j}}_{im}\Vert\e_j-\n_m\Vert).
    \end{split}\label{eq:orbitals}
\end{align}
Here, the exponential envelope from \citet{spencerBetterFasterFermionic2020} guarantees that our wave function will have a finite integral.
In Appendix~\ref{app:wave_function}, we describe how we restrict the envelope parameters $\sigma$ such that we fulfill size consistency desiderata.

\subsection{Optimization}\label{ref:optimization}
We train the whole network end-to-end in a two-step procedure.
We first pretrain the orbitals $\phi_i$ on HF solutions and, next, perform variational optimization~\citep{pfauInitioSolutionManyelectron2020}.

\textbf{Pretraining.} Pretraining is important to ensure a stable variational optimization~\citep{pfauInitioSolutionManyelectron2020,vonglehnSelfAttentionAnsatzAbinitio2023}.
Traditionally, one would match the neural network's orbitals $\phi_i$ with those of an HF solution $\phi^\text{HF}_i$.
But, with our localized orbitals this may cause a mismatch between nuclei involved in the $i$th neural orbital function and HF orbital function as the HF solution is typically sorted by energy state rather than locality.
We resolve this issue by noticing that the HF wave function does not change if one multiplies the orbital matrix $\Phi=\varPhi\Omega^T$ by a matrix $\mA$ with unit determinant, i.e., $\psi(\evec)=\det (\varPhi\Omega^T) = \det (\varPhi\Omega^T\mA)$.
With this in mind, we can find a matrix $\mA$ such that the new coefficient matrix $\hat{\Omega}^T=\Omega^T\mA$ enforces locality.
We describe this optimization procedure in Appendix~\ref{app:canon}.
After finding $\hat{\Omega}^T$, we perform traditional pretraining by matching the neural network orbitals to the localized HF orbitals.
To avoid overfitting, we add a regularization loss on the outputs of the reparametrization network detailed in Appendix~\ref{app:regularization}.

\textbf{Variational optimization.} Like \citet{gaoAbInitioPotentialEnergy2022}, we train both the wave function and the reparametrization network end-to-end and precondition the VMC gradients with natural gradient descent.
But, since we deal with molecules of varying sizes unlike previous work, the gradients obtained from the different molecules may vary by orders of magnitude (like their energy).
To avoid larger molecules from dominating the gradients, we rescale the gradients based on the standard deviation of the local energies associated with the molecule.
We discuss this rescaling in Appendix~\ref{app:rescaling} and the full VMC optimization in Appendix~\ref{app:vmc}.

\subsection{Limitations}
While \our{} can learn a generalized wave function across different geometries of molecules, there are limitations.
Firstly, despite having a rotation equivariant wave function, \our{} is not smooth under arbitrary geometric perturbations.
For instance, changes that cause the equivariant frame from \citet{gaoAbInitioPotentialEnergy2022} to flip result in discrete changes in the wave function, similar discontinuities may happen to our orbital localization as discussed in Appendix~\ref{app:algorithm}.
The coordinate frame also breaks the size-consistency of \ourwf{} as the frames of the two molecules do not necessarily align anymore.
These are general issues introduced by natural symmetries that one must break to model arbitrary wave functions.
Secondly, while we found our gradient rescaling based on the energy's standard deviation easing optimization, we found it to be insufficient if the discrepancy between molecules is large.
For instance, if one trains a hydrogen-based system jointly with heavier atoms like nitrogen we found the optimization resulting in worse results than one obtains from training solely on the hydrogen-based system.
Lastly, in its current state, the number of electrons is equal to the sum of atomic charges, and the number of spin-up and down electrons may differ by at most 1, prohibiting modeling ionic systems.

\section{Experiments}\label{sec:experiments}
Here, we analyze \our{} and \ourwf{} across a variety of different experimental settings.
Firstly, we investigate the behavior of \our{} when training on similar geometries where one would expect significant information overlap between molecular orbitals.
Next, we take a look at its extrapolation behavior on such similar structures.
Thirdly, we train \our{} on molecules that share no common structure.
Fourthly, we investigate the transferability of a trained \our{} to similar and larger molecules.
Lastly, we compare \ourwf{} with recent neural wave functions on the larger benzene molecule.

As the true ground-state energies for any molecular system are rarely known, we either compare them to highly accurate reference calculations or report variational energies or their standard deviation.
As discussed in Section~\ref{sec:background}, VMC energies are upper bounds to the true energy and, thus, lower is better.
Further, as the wave function approaches the ground state, the standard deviation of the local energy approaches zero providing a proxy for the convergence to the ground state.
Appendix~\ref{app:setup} details the setup and Appendix~\ref{app:molecules} lists the geometries we used in the following.
Timings and model sizes can be found in Appendix~\ref{app:time} and Appendix~\ref{app:parameters}.
Appendix~\ref{app:ablation} provides a model size ablation.

\begin{figure}
    \includegraphics[width=\linewidth]{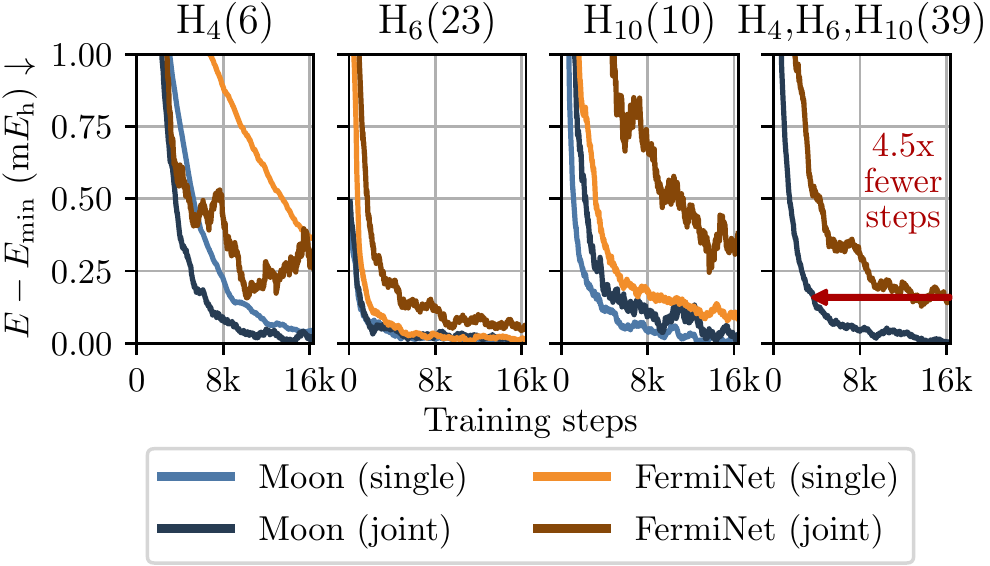}
    \vspace{-0.25in}
    \caption{
        Convergence plots of \our{} with \ourwf{} and FermiNet.
        Numbers in brackets show the number of geometries per molecule.
        In joint training, \ourwf{} converges 4.5 times faster and to lower energies.
    }
    \label{fig:comp_ferminet}
\end{figure}
\begin{figure}
    \centering
    \includegraphics[width=\linewidth]{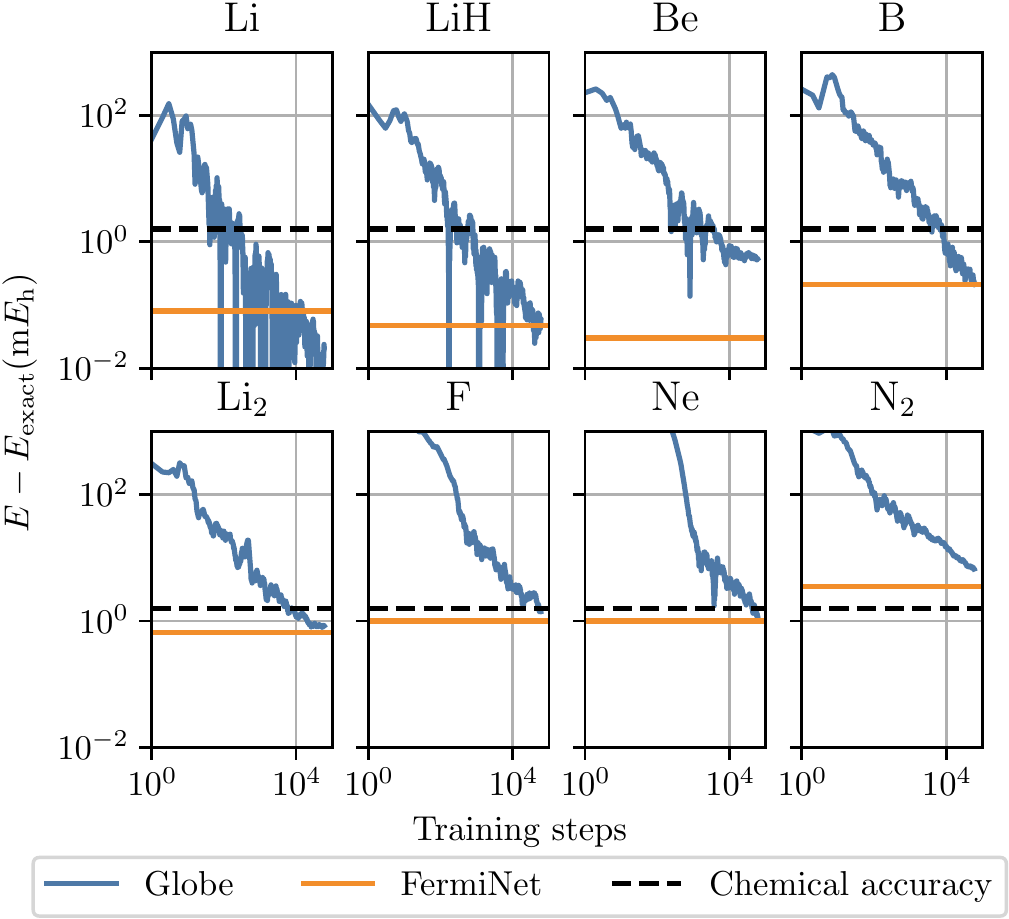}
    \vspace{-0.25in}
    \caption{
        \our{} trained on different structures jointly.
        FermiNet is optimized per molecule,  numbers are taken from \citet{pfauInitioSolutionManyelectron2020}.
        \our{} reaches chemical accuracy everywhere except N$_2$ where the specially optimized FermiNet also fails.
    }
    \label{fig:diverse}
\end{figure}
\begin{figure}
    \includegraphics[width=\linewidth]{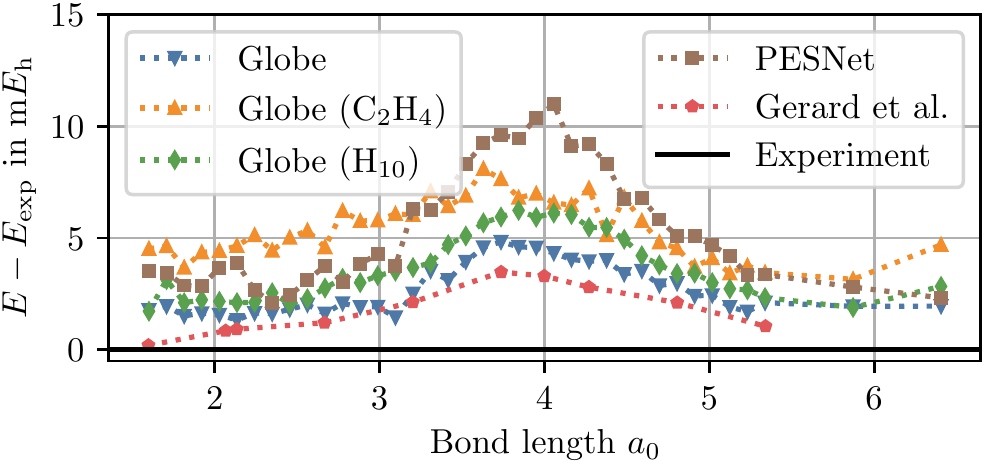}
    \vspace{-0.25in}
    \caption{
        Relative error to \citet{leroyAccurateAnalyticPotential2006} on the N$_2$ potential energy surface.
        For \our{}, the brackets show the additional molecule that the network has trained on.
        The hydrogen chain H$_{10}$ and ethene result in \SI{0.96}{\milli\hartree{}} and \SI{2.71}{\milli\hartree{}} higher energies.
    }
    \label{fig:n2}
\end{figure}
\begin{figure*}
    \includegraphics[width=\linewidth]{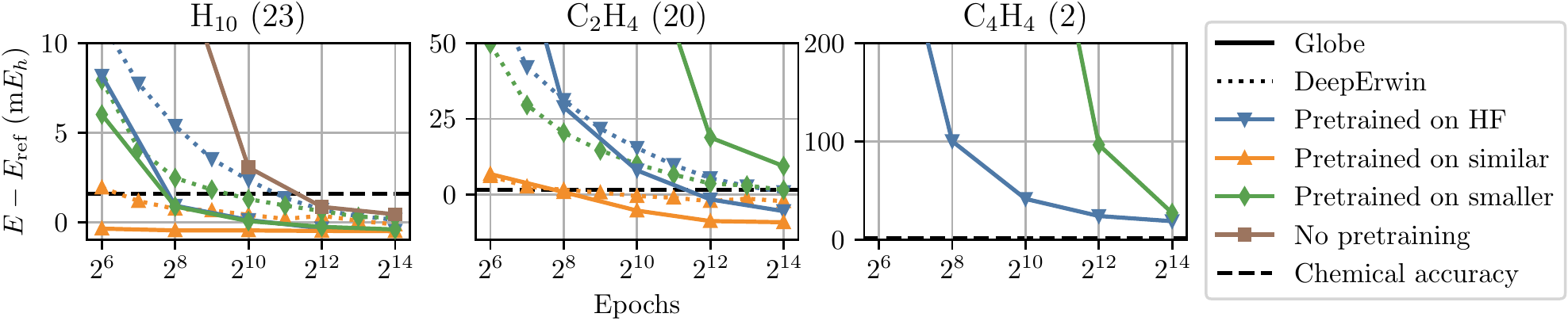}
    \vspace{-0.25in}
    \caption{
        Average energy difference to reference data by training epochs for \our{} and DeepErwin across three molecules with different pretraining techniques.
        Bracketed numbers show the number of geometries.
        For cyclobutadiene, we neither have DeepErwin reference numbers nor did we run the transferability experiment from similar structures.
        On ethene and cyclobutadiene, models without any pretraining are not visible due to their high error, see Appendix~\ref{app:transfer}.
    }\label{fig:transferibility}
    \vspace{-0.15in}
\end{figure*}
\textbf{Learning on similar systems.}
Training on similar systems jointly may be useful in several settings, e.g., in binding energy computations~\citep{trogoloGasPhaseOzoneReactions2019} or in labeling a diverse dataset~\citep{hojaQM7XComprehensiveDataset2021}.
Two key aspects are of interest here.
Firstly, we investigate the loss in accuracy when training on different molecules and, secondly, we compare \ourwf{} to the existing FermiNet~\citep{pfauInitioSolutionManyelectron2020} to identify potential benefits in convergence and accuracy.
For FermiNet, we include all improvements from \citet{gaoSamplingfreeInferenceAbInitio2023} and the Jastrow factor from \citet{vonglehnSelfAttentionAnsatzAbinitio2023}.
As systems, we choose the hydrogen rectangle H$_4$, the 6-element hydrogen chain H$_6$, and the 10-element hydrogen chain H$_{10}$.
Note that each of these systems is a collection of different geometries.
We train each wave function model with our reparametrization network on four different settings, one for each molecule and one where we train on all of them jointly. Due to the variational principle, lower energies are better.

\Figref{fig:comp_ferminet} shows the average energy during training for both FermiNet and \ourwf{} in their two training settings (joint optimization, single molecule only).
One can see that \ourwf{} converges strictly faster than a similar-sized FermiNet.
For small systems, we observe that joint training even accelerates convergence in terms of steps.
For FermiNet, we observe a significant gap between single and joint training for the larger hydrogen chain while \ourwf{} converges to the same energy in both regimes.
Importantly, we observe that \ourwf{} behaves significantly more stable in joint optimization compared to FermiNet.
We compare the standard deviation of the energy during training in Appendix~\ref{app:hydrogen_std}.

\begin{table}
    \caption{
        Comparison between \ourwf{} and FermiNet in solving separated hydrogen chains. H$_{10}$ indicates the energy of a single hydrogen chain, H$_{10}$ + H$_{10}$ for two hydrogen chains \SI{100}{\bohr} apart, and $\Delta$ indicates the difference in energy between twice the energy of single chain compared to solving both systems jointly.
        All values are in Hartree.
    }\label{tab:exten}
    \centering
    \begin{tabular}{lccccc}
        \toprule
        & H$_{10}$ & H$_{10}$ + H$_{10}$ & 2 $\times$ H$_{10}$ & $\Delta$\\
        \midrule
        FermiNet & -5.6631 & -10.7564 & -11.3262 & -0.5698 \\
        \ourwf{} & -5.6632 & -11.3264 & -11.3264 & 0.0000 \\
        \bottomrule
    \end{tabular}
    \vspace{-0.15in}
\end{table}
\textbf{Size consistency.}
As discussed in Desiderata 2 in Section~\ref{sec:method}, for distant molecules the joint wave function is equal to the product of the individual wave functions.
In contrast to previous neural wave functions like FermiNet, \our{} and \ourwf{} adhere to this limit case.
We demonstrate this by training \our{} with FermiNet and \ourwf{} on the hydrogen chain H$_{10}$ and transfer the wave function to two separated hydrogen chains \SI{100}{\bohr} apart.
As an optimal result, one expects the energy of the distant hydrogen chains to be equal to twice the energy of a single chain. 

In Table~\ref{tab:exten} we list the difference between these settings.
While FermiNet leads to a significant error of \SI{570}{\milli\hartree}, \ourwf{}'s result is in perfect agreement with the desiderata.
In closer regimes, we analyze the extensivity of FermiNet and \ourwf{} in Appendix~\ref{app:extensivity}.

\textbf{Learning on dissimilar systems.}
While optimizing similar molecules has a small impact on the performance of \our, we now analyze how unrelated molecules affect final energies.
We test this by solving for the ground-state energies of various small systems jointly.
We pick Li, LiH, Be, B, Li$_2$, F, Ne, and N$_2$ from \citet{pfauInitioSolutionManyelectron2020} as the dataset.

In Figure~\ref{fig:diverse}, one sees the energy for each system during training.
While, for small systems, \our{} with \ourwf{} accomplishes lower energies than FermiNet despite training jointly, we observe that this does not carry over to larger systems where the gap between both increases.
For all molecules, except for N$_2$ where FermiNet also fails to reach chemical accuracy, \our{}'s energies are within the chemical accuracy of the true energy.
To close the gap to FermiNet, we found that increasing the size of the wave function \ourwf{} has a significant impact on the final energy.
We investigate this in Appendix~\ref{app:ablation} and argue that this is due to the increased capacity requirement in capturing many wave functions within a single model.

\textbf{Learning dissimilar energy surfaces.}
While the previous paragraph analyzed the absolute energies in training on different molecules, here we look at the consistency of energy surfaces when training on different energy surfaces jointly.
To test this, we train three \our{} on the nitrogen dimer (N$_2$), once solely on nitrogen, once with the hydrogen chain (H$_{10}$) as a smaller molecule, and once with ethane (C$_2$H$_4$) as a larger molecule.
We chose the nitrogen dimer due to its challenging nature~\citep{pfauInitioSolutionManyelectron2020}.
Further, we compare \our{} to recent neural network-based solutions~\citep{gerardGoldstandardSolutionsSchrodinger2022,gaoAbInitioPotentialEnergy2022}.

The potential energy surface in Figure~\ref{fig:n2} shows that \our{} comes close to the performance of current state-of-the-art neural wave functions while being able to model different molecules jointly.
We suspect the gap to \citet{gerardGoldstandardSolutionsSchrodinger2022} is due to the lower number of determinants, as we use 16 instead of 32 since \citet{pfauInitioSolutionManyelectron2020} found these to be an important hyperparameter for the nitrogen dimer. 
Though, one can see that adding unrelated molecules to the optimization worsens the final results depending on the system sizes.
For instance, adding the relatively low-energy hydrogen chain worsens the results by \SI{0.96}{\milli\hartree}, and the larger ethene structures lead to \SI{2.71}{\milli\hartree} worse energies on average.
Note that the models trained with dissimilar structures also have seen fewer nitrogen samples during training as we evenly divide the total batch size of 4096 across all molecules.

\textbf{Transferibility.}
Here, we analyze the transferability of \our{} across different molecules.
Like \citet{scherbelaSolvingElectronicSchrodinger2022}, we train \our{} on several molecules and transfer the wave function either to different geometries, or larger molecules.
Specifically, we use the 10-element hydrogen chain (H$_{10}$), ethene (C$_2$H$_4$), and cyclobutadiene (C$_4$H$_4$).
As smaller molecules, we use the 6-element hydrogen chain (H$_{6}$), methane (CH$_4$), and ethene, respectively.
We compare our results to DeepErwin~\citep{scherbelaSolvingElectronicSchrodinger2022}.
Though, there are key distinctions in our setups.
While \our{} optimizes all molecules at once, DeepErwin optimizes each molecule independently with weight-sharing applied between the wave functions.
Further, DeepErwin performs new CASSCF calculations for each molecule while we apply \our{} without any SCF calculations to the new molecule.
An epoch for DeepErwin is one optimization step per molecule with a batch size of 2048 per molecule.
Since we optimize all geometries jointly, an epoch is a single step for us with a batch size of 4096 shared for all molecules in the batch. 
Thus,  we trained \our{} with approximately 10 times fewer samples and 20 times fewer steps.

The convergence plots in \Figref{fig:transferibility} show that \our{} generally converges quickly to within chemical accuracy in the hydrogen chain and ethene for the standard setting, i.e., pretraining from HF.
For cyclobutadiene, we suspect the gap to \citet{gaoSamplingfreeInferenceAbInitio2023} is due to our smaller wave function, i.e., we use an embedding dim of 256 and 16 determinants instead of 512 and 32, respectively. 
Consistent with the results from \citet{scherbelaSolvingElectronicSchrodinger2022}, we find starting from smaller molecules generally worsens convergence for larger systems.
Still, pretraining on smaller molecules leads to similar results given sufficient training steps and outperforms the models without pretraining.
For a view of the full convergence diagram, see Appendix~\ref{app:transfer}.
Notably, we observe convergence without performing a single HF calculation on the larger structures, which was essential for previous methods~\cite{pfauInitioSolutionManyelectron2020,scherbelaSolvingElectronicSchrodinger2022}.
Compared to DeepErwin, \our{} results in \SI{0.6}{\milli\hartree} and \SI{6.3}{\milli\hartree} lower energies for the hydrogen chain and ethene, respectively.

\begin{figure}
    \centering
    \includegraphics[width=\linewidth]{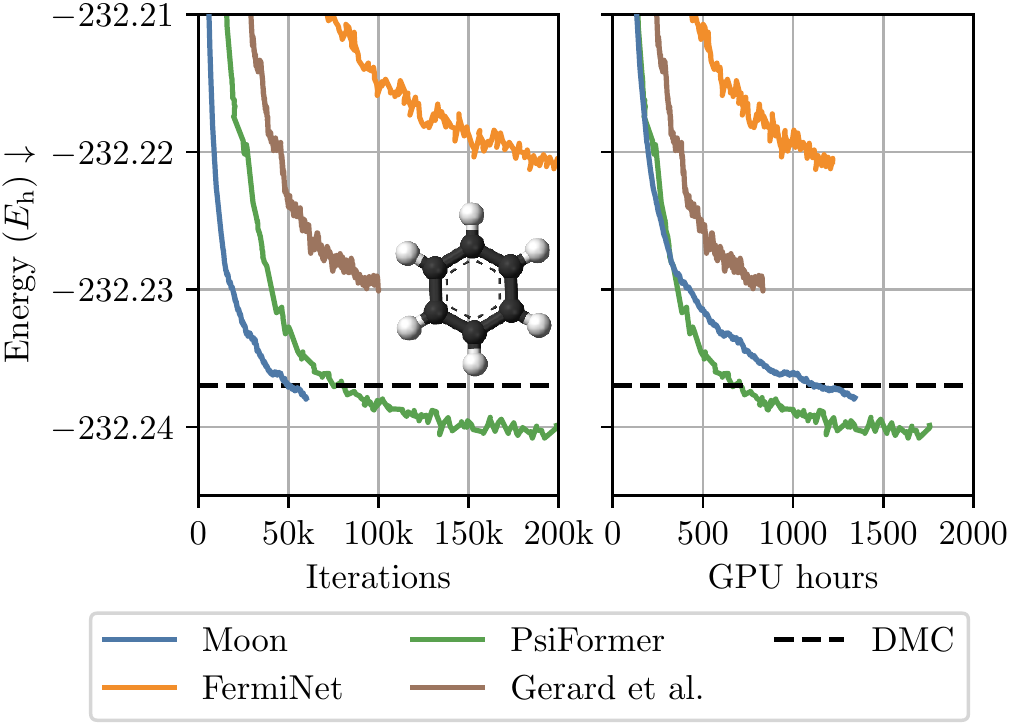}
    \vspace{-0.25in}
    \caption{
        Energy during optimization on benzene (C$_6$H$_6$) by iterations and GPU hours. Reference data is taken from \citet{vonglehnSelfAttentionAnsatzAbinitio2023,gerardGoldstandardSolutionsSchrodinger2022,renGroundStateMolecules2022}. Energies are averaged over 4000 iterations.
        \ourwf{} shows similar convergence behavior to the attention-based PsiFormer.
    }
    \label{fig:benzene}
    \vspace{-0.15in}
\end{figure}

\textbf{Larger systems.} 
As neural-network solutions are interesting thanks to their theoretical scaling, we investigate how \ourwf{} scales to larger systems compared to FermiNet~\citep{pfauInitioSolutionManyelectron2020}, recent improvements to FermiNet~\citep{gerardGoldstandardSolutionsSchrodinger2022}, diffusion Monte Carlo (DMC) calculations~\citep{renGroundStateMolecules2022}, and the recently proposed attention-based PsiFormer~\citep{vonglehnSelfAttentionAnsatzAbinitio2023}.
As system, we use the benzene molecule (C$_6$H$_6$), as recent work found a large gap between VMC and the true ground-state energy~\citep{renGroundStateMolecules2022,vonglehnSelfAttentionAnsatzAbinitio2023}.

Figure~\ref{fig:benzene} plots the energy of the system throughout training by iterations and GPU hours.
Note that FermiNet, PsiFormer, and \citet{gerardGoldstandardSolutionsSchrodinger2022} were optimized with KFAC~\citep{martensOptimizingNeuralNetworks2015} while \ourwf{} uses CG-based natural gradient descent.
While CG-based natural gradient descent leads to faster early convergence, KFAC results in similar update steps later in training while being two to three times faster per iteration.
In energy, we find \ourwf{} to behave similarly to PsiFormer in that it approaches lower than DMC energy levels. 
Compared to the best FermiNet results, we find \ourwf{} to reach similar energies in 3 times fewer GPU hours and converge in an identical time to \SI{13.5}{\milli\hartree} lower energies.

\section{Conclusion}\label{sec:discussion}
Solving many Schrödingers jointly with a single system holds the promise of learning generalizing wave functions.
Like recent deep learning-based density functional theory (DFT) functionals~\citep{snyderFindingDensityFunctionals2012,kirkpatrickPushingFrontiersDensity2021}, learning a general neural network solution for quantum mechanical calculations may accelerate material discovery while increasing accuracy.
In this work, we introduced \our{}, the first systematic approach to performing such a generalization.
By embedding orbitals as points in space and using message passing, we can learn dynamic numbers of molecular orbital functions for different molecules.
Further, we presented a novel locality-driven wave function, \ourwf{}, that shows significant improvements in convergence and extensivity than previous methods when trained on diverse molecules.
With \our{} and \ourwf{}, we are the first to demonstrate the solving of Schrödinger equations of different molecules within a single wave function.

\textbf{Acknowledgements.}
We greatly thank Chendi Qian for his valuable early work on designing wave functions and Michael Scherbela for swiftly sharing their data and results with us.
Further, we would like to thank Tom Wollschläger and Marten Lienen for their discussion on the localization algorithm, and Jan Schuchardt and Yan Scholten for their invaluable feedback on the paper.

Funded by the Federal Ministry of Education and Research (BMBF) and the Free State of Bavaria under the Excellence Strategy of the Federal Government and the Länder.

{
\bibliography{../mabenet}
\bibliographystyle{icml2023}
}
\clearpage
\appendix
\section{Size consistency in quantum chemistry}\label{app:size_consistency}
If one is given two distant molecules, one can show that the Hamiltonian $\mH$ from Equation~\ref{eq:schroedinger} decomposes into two Hamiltonians $\mH_1,\mH_2$ for each of the respective systems.

W.l.o.g., let the electrons be sorted such that $\evec=[\evec_1,\evec_2]\in\R^N, \evec_1\in\R^{N_1}, \evec_2\in\R^{N_2}$.
First, one rewrites the full Hamiltonian
\begin{align}
    \begin{split}
        \mH =& -\frac{1}{2}\sum_{n=1}^{3N}\nabla_n^2 \\
            &+ \sum_{n>m=1}^N\frac{1}{\Vert r_n - r_m\Vert}
            - \sum_{n=1}^N\sum_{m=1}^M \frac{Z_m}{\Vert r_n - R_m\Vert} \\
            &+\sum_{m>n=1}^M \frac{Z_mZ_n}{\Vert R_m - R_n\Vert},
    \end{split}
\end{align}
in terms of the Hamiltonians of the individual systems $\mH_1,\mH_2$
\begin{align}
    \begin{split}
        \mH =& \mH_1 + \mH_2
            + \sum_{n\in\sB_1}\sum_{m\in\sB_2} \frac{1}{\Vert\e_n-\e_m\Vert}\\
            &- \sum_{n\in\sB_1}\sum_{m\in\sA_2} \frac{Z_m}{\Vert\e_n-\n_m\Vert}\\
            &- \sum_{n\in\sB_2}\sum_{m\in\sA_1} \frac{Z_m}{\Vert\e_n-\n_m\Vert}\\
            &+ \sum_{n\in\sA_1}\sum_{m\in\sA_2} \frac{Z_mZ_n}{\Vert\e_n-\n_m\Vert}.
    \end{split}
\end{align}
where $\sA_1,\sA_2$ and $\sB_1,\sB_2$ are the index sets for the nuclei and electrons for both systems, respectively.
For distant systems, $\frac{1}{\Vert\e_n-\e_m\Vert}\approx \frac{1}{\Vert\e_n-\n_m\Vert}\approx\frac{1}{\Vert\n_n-\n_m\Vert}$:
\begin{align}
    \begin{split}
        \mH =& \mH_1 + \mH_2
            +  c(\\
            &\underbrace{\phantom{-} \sum_{n\in\sB_1}\sum_{m\in\sB_2} 1
            - \sum_{n\in\sB_1}\sum_{m\in\sA_2} Z_m}_{=0}\\
            &\underbrace{- \sum_{n\in\sB_2}\sum_{m\in\sA_1} Z_m
            + \sum_{n\in\sA_1}\sum_{m\in\sA_2} Z_mZ_n}_{=0}).
    \end{split}
\end{align}
where $c=\frac{1}{\Vert\n_n -\n_m\Vert}, n\in \sA_1, m\in\sA_2$. One may notice that for non-ionic systems $\vert\sB_i\vert=\sum_{m\in\sA_i}Z_m$. Thus, the first two and the last two sums cancel out and one is left with the sum of the individual Hamiltonians.

Given the decomposition of the Hamiltonian, one can show that the lowest eigenvalue of $\mH$ is $E_1 + E_2$ where $E_1$ and $E_2$ are the lowest eigenvalues associated with the eigenfunctions $\psi_1$ and $\psi_2$ of $\mH_1$ and $\mH_2$, respectively:
\begin{align}
    \mH(\psi_1\psi_2) &= (\mH_1 + \mH_2)(\psi_1\psi_2) \\
    &= \mH_1\psi_1\psi_2 + \mH_2\psi_2\psi_1 \\
    &= E_1\psi_1\psi_2 + E_2\psi_2\psi_1 \\
    &= (E_1 + E_2)\psi_1\psi_2.
\end{align}
Thus, the ground state wave function of the combined Hamiltonian $\mH$ is the product of the individual ground state wave functions $\psi_1$ and $\psi_2$.

To obtain the decomposition $\psi=\psi_1\psi_2$, it remains to show that the molecular orbital functions $\phi$ must only act on close-by electrons, i.e.,
\begin{align}
    \phi_n(\e_m\vert \evec) = \begin{cases}
        \phi_n(\e_m\vert \evec_{\sB_1}) & \text{if } n,m\in\sB_1, \\
        \phi_n(\e_m\vert \evec_{\sB_2}) & \text{if } n,m\in\sB_2, \\
        0 & \text{else.}
    \end{cases}. \label{eq:local_orbital}
\end{align}
where we introduced the shorthand notation $\evec_{\sB_i}=\{\e_n\}_{n\in\sB_i}$.
For simplicity, we assume a single Slater determinant as wave function $\psi$:
\begin{align}
    \psi(\evec) = \det\begin{pmatrix}
        \phi_1(\e_1\vert \evec) & \hdots & \phi_N(\e_1\vert \evec)\\
        \vdots & \ddots & \vdots\\
        \phi_1(\e_N\vert \evec) & \hdots & \phi_N(\e_N\vert \evec)
    \end{pmatrix}.
\end{align}
If we plug in the definition from Equation~\ref{eq:local_orbital}, the matrix factorizes into a block diagonal which then in turn factorizes into the product of wave functions as desired:
\begin{align}
    \psi(\evec) =& \det\begin{pmatrix}
        \phi_1(\e_1\vert \evec_{\sB_1}) & \hdots & 0\\
        \vdots & \ddots & \vdots\\
        0 & \hdots & \phi_N(\e_N\vert \evec_{\sB_2})
    \end{pmatrix}\\
    =& \det \Phi_1(\evec_{\sB_1}) \det \Phi_2(\evec_{\sB_2})\\
    =&\psi_1(\evec_{\sB_1})\psi_2(\evec_{\sB_2}).
\end{align}

\section{Orbital localization algorithm}\label{app:algorithm}
\begin{algorithm}[tb]
    \caption{Orbital localization}
    \label{alg:orbital_localization}
    \begin{algorithmic}
        \STATE {\bfseries Input:} nuclei positions $\n_i\in\R^3$, charges $Z_i\in\mathbb{N}_+$
        \STATE Orbital locations $Locs=[\phantom{a}]$
        \STATE Orbital types $Types=[\phantom{a}]$
        \STATE \textit{\# Define core orbitals}
        \FOR {$i=1; i\leq M$}
            \STATE Valence orbitals $V_i := \text{Valency}(Z_i)$
            \STATE Core orbitals $C_i := \ceil{\frac{Z_i - V_i}{2}}$
            \FOR {$j=1; j \leq C_m$}
                \STATE $Locs$.append($R_i$)
                \STATE $Types$.append($(Z_i, j)$)
            \ENDFOR
        \ENDFOR
        \STATE Distances $D_{m,n} := \begin{cases}
            \Vert R_m - R_n\Vert &\text{, if } n \neq m\\
            c_\text{self}&\text{, else}
        \end{cases}$
        \STATE \textit{\# Define valence orbitals}
        \STATE Bond type $T_{m,n} := 0$
        \FOR {$i=1; i\leq\ceil{\sum_{m}^M V_m/2}$}
            \STATE Scores $S_{m, n} := \frac{\1[V_m > 0 \land V_n > 0]}{D_{m, n} + T_{m, n}/2}$
            \STATE Indices $m, n := \argmax_{m,n} \mS$ (w.l.o.g. $m\leq n$)
            \STATE $V_m := V_m - 1; V_n := V_n - 1$
            \STATE $T_{m,n} := T_{n,m} := T_{m,n} + 1$
            \STATE $Locs$.append($\frac{R_m+R_n}{2}$)
            \STATE $Types$.append($T_{m, n}$)
        \ENDFOR
        \STATE \textbf{return} $Locs, Types$
    \end{algorithmic}
\end{algorithm}
In an ideal setting, we would pick the orbital locations such that a) the \emph{local} environment defines them, b) they are \emph{deterministic} and c) they change \emph{smoothly} with arbitrary changes to the atoms.
Except for a few edge cases, which we discuss in the next paragraph, we satisfy all three criteria with Algorithm~\ref{alg:orbital_localization}.
As explained in Section~\ref{sec:reparam}, we distinguish between core and valence orbitals.
Where core orbitals are located at their corresponding nucleus, valence orbitals are located at the center of covalent bonds.
To favor the formation of diverse spatially different bonds, higher bond types (e.g., double and triple bonds) are slightly punished such that one favors similar distanced single bonds.
Since the distance of an atom to itself is always 0, we replace the self-distances by a cutoff radius $c_\text{self}$ after which one prefers self-bonds.

\textbf{Edge cases.}
While the orbital localization fulfills our free desiderata: locality, determinism, and smoothness most of the time, there are edge cases we would like to highlight here in which we cannot guarantee smoothness.
First, discrete changes occur when the nearest neighbors between atoms change.
Second, if multiple pairs have identical distances, the algorithm would not be deterministic.
In such cases, we rely on a series of `tie-breakers', i.e., further criteria.
In particular, we prefer edges furthest from the center of mass.
If a tie remains, we start comparing the polar angle and, lastly, the azimuthal angle to break ties.
While this formulation allows us to localize orbitals deterministically it also breaks the smoothness, e.g., if the order in any of the tie-breakers changes. 
Further, it relies on a smoothly changing equivariant coordinate frame which cannot exist~\citep{gaoAbInitioPotentialEnergy2022}.

Considering these discrete jumps in our orbital localization, one may ask why we decided on this particular algorithm.
To answer this, one first has to consider why these discrete changes happen within \Algref{alg:orbital_localization}.
The $\argmax$ function introduces these discrete changes.
While replacing the $\argmax$ by a smooth approximation, e.g., via a softmax would resolve all discrete jumps, it would greatly deteriorate the locality.
For instance, in larger systems, the softmax will always surely converge to the center of mass rather than any local bond structure.
The examples given above are in extreme situations in symmetric molecules, e.g., the transition state of cyclobutadiene.
In our experience, even established classical approximative methods such as the Hartree-Fock method struggle in such situations.
One may see localizing orbitals as an instantiation of the greater problem of how one should break symmetries in neural wave functions such that one can model the ground state accurately while keeping sufficient inductive bias to generalize to new structures.

As a final point of discussion, one should ask whether these discontinuities are harmful in practice.
As for equilibrium structures, the cases we listed will generally not happen as every atom will be closely surrounded by as many atoms as its valency with longer distances to other atoms.
Considering these aspects, we believe our orbital localization algorithm to be sufficient for the current state of neural wave functions while we encourage future work to approach the problem of discontinuities.

\section{\ourwflong{} details}\label{app:wave_function}
As Section~\ref{sec:our_wf} focuses on novel aspects of the wave function, we want to provide some implementation and minor details here.

\textbf{Rescaling.}
To limit the input magnitude for distanced particles, we adopt the logarithmic rescaling from \citet{vonglehnSelfAttentionAnsatzAbinitio2023}, i.e.,
\begin{align}
    \vg_{ij}=\frac{\log (1+g_{ij}^{(4)})}{g_{ij}^{(4)}} \vg_{ij}.
\end{align}

\textbf{Normalization.}
Like the reparametrization network, we use learnable normalization factors within the wave functions for the four message-passing steps \Eqref{eq:elec_init}, (\ref{eq:elec_emb}), (\ref{eq:nuc_emb}) and (\ref{eq:diff}).
Specifically, we normalize the electron embeddings after the electron-electron message passing in \Eqref{eq:elec_init} by the expected number of close electrons via
\begin{align}
    \hat{\vh}^{\text{e} (0)}_i &= \frac{1}{\mu(\e_i)} \vh^{\text{e} (0)},\\
    \mu(\e) &= 1 + \sum_{m=1}^M \frac{Z_m}{2} \exp\left(-\frac{\Vert \e-\n_m\Vert^2}{\sigma^2_\text{norm}}\right)
\end{align}
where $\hat{\vh}^{\text{e} (0)}_i$ are the electron embeddings passed to further layers. Note that this formulation is similar to \Eqref{eq:norm} but here we multiply by half of the charge of the nucleus to account for the expected number of electrons per spin close to the nucleus.
For the electron-nuclei message-passing steps in \Eqref{eq:elec_emb}, (\ref{eq:nuc_emb}) and (\ref{eq:diff}), we use $\frac{1}{\nu_{\e_i}^\nvec}$, $\frac{1}{\nu_{\n_i}^\nvec}$, and $\frac{1}{\nu_{\e_i}^\nvec}$, respectively.

\textbf{Reparametrized filters.}
In the message-passing steps in \Eqref{eq:elec_emb}, (\ref{eq:nuc_emb}) and (\ref{eq:diff}), we use reparametrized version of the spatial filters from \Eqref{eq:filter}.
Concretely, these reparametrized versions take the form
\begin{align}
    \bar{\filter}^{(l)}_m(\vx) =& \mW^{(l)}\bar{\beta}_m(\vx),\\
    \begin{split}
        \bar{\beta}_m(\vx) =& \mW^\text{env} \left[ \exp\left(-\left(\frac{\Vert \vx\Vert}{\bar{\varsigma}_{mi}}\right)^2\right) \right]_{i=1}^{D}\\
        &\circ \left(\act\left(\vx\bar{\mW}_m^{(1)}+\bar{\vb}_m^{(1)}\right)\mW^{(2)} + \vb^{(2)}\right). \label{eq:adapt_filter_beta}
    \end{split}
\end{align}
Here, we replaced the envelope ranges $\varsigma_i$ and the first linear layer in the MLP by atom-parameterized versions.
The remaining parameters are shared across all $m$.

\textbf{Residual connections.} We add residual connections between each update layer and renormalize the embeddings, i.e.,
\begin{align}
    \hat{\vh}^{\text{n} (L+1)}_m = \frac{1}{\sqrt{2}} \left(
        \vh^{\text{n} (L)}_m
        + \vh^{\text{n} (L+1)}_m
    \right)
\end{align}
where $\hat{\vh}^{\text{n} (L+1)}_m $ are the nuclei embeddings used in subsequent layers.
For the electron embeddings, we add a skip connection after the diffusion step
\begin{align}
    \hat{\vh}^{\text{e} (L)}_i = \frac{1}{\sqrt{2}}
    \left(
        \vh^{\text{e} (L)}_i + \vh^{\text{e} (0)}_i
    \right).
\end{align}

\textbf{Jastrow factor.} As Jastrow factor we additively combine the Jastrow factors from~\citet{gaoSamplingfreeInferenceAbInitio2023} and \citet{vonglehnSelfAttentionAnsatzAbinitio2023}
\begin{align}
    \begin{split}
        J(\evec) =& \sum_{i=1}^{N}\text{MLP}(\vh^{\text{e} (L)}_i) \\
        &+ \beta_\text{par}\sum_{i,j;\spin_i=\spin_j} -\frac{1}{4}\frac{\spin_\text{par}^2}{\spin_\text{par} + \Vert\e_i-\e_j\Vert}\\
        &+ \beta_\text{anti}\sum_{i,j;\spin_i\neq\spin_j} -\frac{1}{2}\frac{\spin_\text{anti}^2}{\spin_\text{anti} +  \Vert\e_i-\e_j\Vert}.
    \end{split}
\end{align}
where $\alpha_\text{par}, \alpha_\text{anti}, \beta_\text{par}, \beta_\text{anti}$ are learnable parameters.

\textbf{Parameter domain.}
While most of the reparametrized parameters are weight matrices without clear restrictions, some are used in numerically critical situations, e.g., as a divisor.
In such cases, we apply a softplus function $f(x)=\log(1+\exp(x))$ to avoid division by zero.
Concretely, we use this domain restriction for the envelope ranges $\tilde{\bar{\varsigma}}_m$ and envelope parameters $\tilde{\bar{\sigma}}$.
To obtain local orbitals, $\tilde{\bar{\pi}}$ decay to zero if the distance between an atom and an orbital increases. 
We accomplish this by defining $\tilde{\bar{\pi}}=\tanh(\tilde{\bar{\pi}}_1)f(\tilde{\bar{\pi}}_1)$ where $\tilde{\bar{\pi}}_1$ and $\tilde{\bar{\pi}}_2$ are two different outputs of the reparametrization network and $f$ is the softplus function.
As any direct atom-orbital parameter decays to 0 if the distance between the atom and the orbital increases, this parametrization gives the desired effect.
While one could also drop any transformation on $\tilde{\bar{\pi}}$ to accomplish the decaying effect, \citet{gaoAbInitioPotentialEnergy2022} found a softplus on $\pi$ to help in convergence which we confirmed in early experiments.
Thus, we define $\tilde{\bar{\pi}}$ as a product where the $\tanh$ accounts for the decay and sign, and the softplus for the granularity.
In Appendix~\ref{app:size_consistency}, we show that this parametrization results in the desired product of wave functions for distant systems.

\section{Canonicalizing Hartree-Fock solutions}\label{app:canon}
\begin{figure}
    \includegraphics[width=\linewidth]{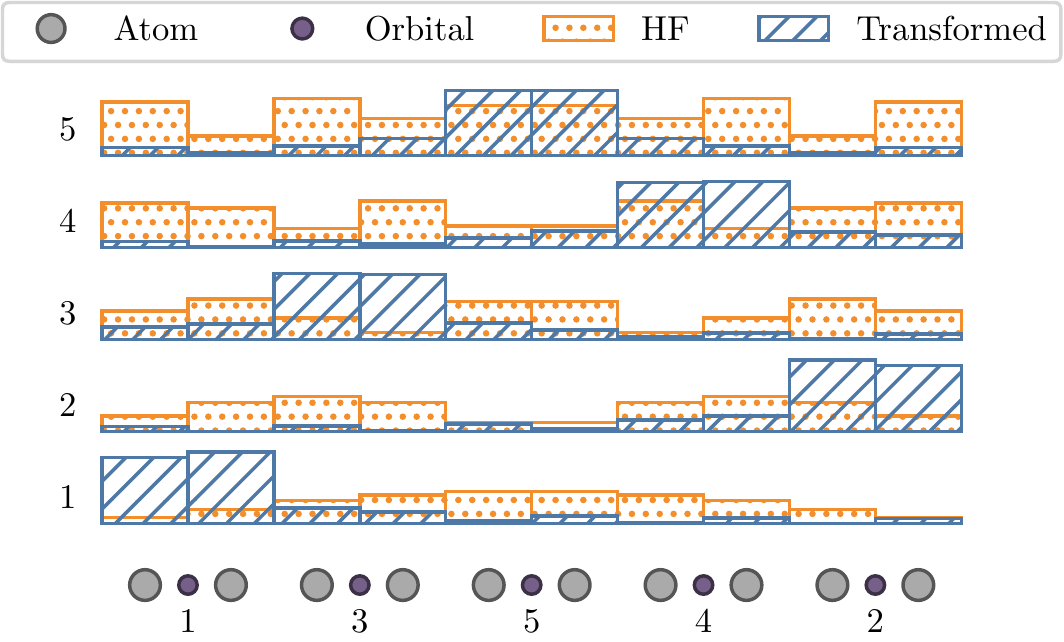}
    \vspace{-0.25in}
    \caption{
        Illustration of our orbital canonicalization.
        The bottom row illustrates the hydrogen chain with our localized orbitals.
        Rows 1 to 5 illustrate the contribution of each atom to each of the five orbitals for the Hartree-Fock solution obtained via PySCF~\citep{sunPySCFPythonbasedSimulations2018} and our transformed solution.
        We plot the sum of absolute values of $\Omega$ belonging to atom $i$ as the height of the $i$th bar.
        Our transformed solution (blue) localizes the contribution close to the nuclei closest to its location.
    }
    \vspace{-0.15in}
    \label{fig:hf_transform}
\end{figure}
\begin{algorithm}[tb]
    \caption{Mask construction}
    \label{alg:mask_construction}
 \begin{algorithmic}
    \STATE {\bfseries Input:} Atom pairs $(n_i, m_i)$, Orbital types $T_i\in\mathbb{N}_+$, Atomic orbitals $O_m$
    \STATE \textit{\# $\sP$ is a dictionary of dictionaries of lists where the first level is an atom, the second level are orbital types, and the list keeps track of the orbitals.}
    \STATE Priority $\sP := \{\}$
    \FOR {$i=1; i\leq N_\text{orb}$}
        \STATE $\sP[n_i][T_i].\text{append}(i)$
        \IF {$n_i \neq m_i$}
            \STATE $\sP[m_i][T_i].\text{append}(i)$
        \ENDIF
    \ENDFOR

    \STATE Offsets $o_i := \sum_{m=1}^{i-1} O_m$
    \STATE Result $\mM := \bm{0}^{\sum_{m=1}^{M}\times N}$

    \STATE \textit{\# Iterate through all atoms.}
    \FOR {$n, \sT_{n} \in \sP$}
        \STATE \textit{\# Offset that indicates the lowest free orbital.}
        \STATE Offset $o := o_n$
        \STATE \textit{\# Iterate through all orbitals types of atom $n$.}
        \FOR {$T,\vm \in \sT_n$}
            \STATE \textit{\# Iterate through the cardinality orbital type $T$.}
            \FOR {$i=1; i\leq \dim(\vm)$}
                \STATE $\mM_{o+i,m_i} := 1$
            \ENDFOR
            \STATE \textit{\# Since $\dim(\vm)$ many orbitals have been assigned, we must increase the offset.}
            \STATE $o := o + \dim(\vm)$
        \ENDFOR
    \ENDFOR
    \STATE \textbf{return} $\mM$
 \end{algorithmic}
 \end{algorithm}
As discussed in Section~\ref{ref:optimization}, the solution obtained from a Hartree-Fock calculation may not align with our assumptions about the locality of orbitals.
Since learning global orbital functions based on our localized orbital embeddings presents a difficult challenge, we seek to canonicalize our Hartree-Fock solutions such that they align with our localized orbitals.
As explained in Section~\ref{sec:background}, in Hartree-Fock one optimizes coefficients of linear combinations of atomic orbital functions to construct molecular orbital functions.
\Figref{fig:hf_transform} shows these coefficients $\Omega^T$.
Each Hartree-Fock molecular orbital, 1 to 5, exhibits a non-local structure.
In the following, we will describe how we transform obtained Hartree-Fock solutions such that they obey a local structure and canonicalize their coefficients to avoid disagreements between similar molecules. 

If one considers the HF electronic wave function $\psi(\evec)=\det\Phi=\det\varPhi\Omega^T$, one can see that the wave function does not change if one multiplies the coefficient matrix $\Omega^T$ with a matrix with unit determinant $\mA\in\R^{N\times N}$ on the right, i.e., $\psi(\evec)=\det\varPhi\Omega^T\mA$.
As the sign of the wave function is arbitrary, $\mA$ may also have a determinant of $-1$.
Since we know the atoms involved in the localization of the $i$th molecular orbital, we can formulate an optimization problem as
\begin{align}
    \min_\mA& \Vert\Omega^T\mA \circ (1-\mM)\Vert^2_2 + \sum_{i=1}^{N} (1 - \Vert (\Omega^T\mA\circ \mM)_i\Vert_2)^2 \label{eq:hf_opt}\\
    \text{s.t.}& \vert\det\mA\vert = 1\label{eq:hf_constr}
\end{align} 
where $\mM \in \{0, 1\}^{N \times \eta}, \eta=\sum_{m=1}^{M} O_m$ is a binary mask indicating our desired relation between atomic and molecular orbitals.
Note, that the first term in \Eqref{eq:hf_opt} encourages zero interaction with non-involved atoms, and the second term aids in keeping the wave function normalized.

In our matching mask $\mM$, we want to preserve the order of energy levels.
Atomic orbitals are typically sorted by energy level, i.e., the $i$th orbital has lower energies than the $j$th atomic orbital iff $i<j$.
To get a canonical ordering, we translate these energy levels to our localized orbitals.
Specifically, we want our localized core orbitals to match the atomic orbitals in the same order, i.e., our $i$th core orbital of an atom should match the $i$th atomic orbital.
For valence orbitals, we enforce the same for bonds of higher order (double bonds, triple bonds, ...) where we match the first valence orbital of that bond to the lowest free orbitals of both atoms and the second bond to the second lowest free orbitals of both, etc.
If an atom has bonds to $k$ different atoms, we cannot easily define an order between bonds and, thus, distribute the $k$ free orbitals of both atoms equally to the $k$ different bonds.

We illustrate this in an example of H$_2$O.
Hydrogen has 1 atomic orbital and oxygen 5 (the number of atomic orbitals is determined by the period of the element), i.e., $\eta=6$.
The number of molecular orbitals is $\ceil{\frac{\sum_{m=1}^{M}Z_m}{2}}=5$.
Now, assuming an equilibrium structure, our molecular orbitals are distributed as follows: three core orbitals associated with oxygen and one valence orbital for each of the O-H bonds.
Given our mask construction, we get the following mask
\begin{align}
    \mM_\text{H2O} &= 
    \mleft[
        \begin{array}{ccccc|c|c}
            1 & 0 & 0 & 0 & 0 & 0 & 0 \\
            0 & 1 & 0 & 0 & 0 & 0 & 0 \\
            0 & 0 & 1 & 0 & 0 & 0 & 0 \\
            \hline
            0 & 0 & 0 & 1 & 1 & 1 & 0 \\
            0 & 0 & 0 & 1 & 1 & 0 & 1 \\
        \end{array}
    \mright]
\end{align}
where the first three rows are core-orbitals and the last two rows are valence-orbitals.
The vertical lines group the atomic orbitals by the associated atom, the first five columns belong to the oxygen atom while the last two belong to each of the hydrogen atoms.
A formal definition of our mask construction is given in \Algref{alg:mask_construction}.

Finally, we solve \Eqref{eq:hf_opt} with an alternating optimization algorithm.
In the first step, we optimize $\mA$ with the Broyden-Fletcher-Goldfarb-Shannon (BFGS) algorithm~\citep{nocedalNumericalOptimization2nd2006} where we parametrize $\hat{\mA}$ with real numbers but normalize it with $\mA = \frac{1}{\sqrt[N]{\vert \det \mA\vert}}\hat{\mA}$ before computing the loss.
Because this restriction cannot change the sign of the determinant $\mA$ and permutation represent local minima, we use the Hungarian algorithm~\citep{kuhnHungarianMethodAssignment1955} to optimize over all permutations given a fixed $\mA$.
We compute the cost matrix for the Hungarian algorithm by evaluating the \Eqref{eq:hf_opt} for all possible pairwise permutations.
After finding the optimal permutation matrix $\mP^{(t)}$, we merge it into $\mA^{(t+1)}=\mA^{(t)}\mP^{(t)}$.
We either stop after a fixed number of iterations or after the loss does not change.
Typically, this method converges within 2 iterations.
Since we have to do this only once as preprocessing before pretraining, one can neglect the computational cost, which is in the order of a second per molecule.

As the sign of the wave function is arbitrary, we should decide on a canonical sign for each molecular orbital to avoid mismatches between different Hartree-Fock solutions of similar structures.
We implement this by multiplying $\mM$ with a diagonal matrix $\mD$ where the diagonal elements are defined as
\begin{align}
    \mD_{ii} = \begin{cases}
        -1 & \text{ if } \sum_{j=1}^{\eta}(\Omega^TA\circ \mM)_{ij} < 0,\\
        1 & \text{else}.
    \end{cases}
\end{align}

\section{Pretraining regularization}\label{app:regularization}
Since we found the output of the reparametrization network to be unstable during pretraining, we add a small regularization to its output.
Specifically, we define for each parameter matrix it outputs a target normal distribution, i.e., a mean and variance.
For instance, for weight matrices $\bar{\mW}\in\R^{d_\text{in}\times d_\text{out}}$, such the ones in \Eqref{eq:adapt_filter_beta} or \Eqref{eq:elec_nuc_pair}, we follow the standard initialization and define the mean to be zero and the standard deviation to be $\frac{1}{\sqrt{d_\text{in}}}$~\citep{lecunEfficientBackprop2012}.
In our regularization loss, we then enforce that the outputted distribution follows our target normal distribution by matching the first $p_\text{max}$ moments of the output distribution with the moments of the target distribution.
Concretely, for the $i$th outputted parameters we add the following loss
\begin{align}
    \mathcal{L}_\text{pre}(\theta_i) =& \sum_{p=1}^{p_\text{max}}\left(\frac{1}{\vert\hat{\theta}_i\vert}\left(\sum_{j=1}^{\vert\hat{\theta}_i\vert}\hat{\theta}_{ij}^p\right) - m_p\right)^2, \\
    \hat{\theta}_i =& \frac{(\theta_i - \mu_i)}{s_i}, \\
    m_p =& \begin{cases}
        0 & \text{if }$p$\text{ is odd}, \\
        (p - 1)!! & \text{if }$p$\text{ is even}
    \end{cases}
\end{align}
where $\theta_i$ is $i$th outputted parameter, $\mu_i$ is its target mean, $s_i$ its target standard deviation, $!!$ the double factorial and $m_p$ is the $p$th central moment of a standard normal distribution.

\section{Rescaling gradients}\label{app:rescaling}
In the following, we discuss a gradient rescaling technique on a per-molecule basis to obtain a stable optimization if one optimizes molecules of different sizes jointly.
As the norm of gradients in \Eqref{eq:gradient} is proportional to the expected deviation from the mean, the standard deviation of the energy functions as a proxy for the gradient's norm.
We rescale the gradients based on this proxy rather than the actual gradient norm as acquiring the latter is inherently expensive as it requires one to compute the full Jacobian of the network rather than a Jacobian vector product.
Given the standard deviations $s_i=\sqrt{\E_{x\sim \psi_{\theta_i}^2}\left[
    E_{\theta_i}(x) - \E_{x\sim \psi_{\theta_i}^2}\left[E_{\theta_i}(x)\right]
\right]^2}$
where $\theta_i$ indicates the parameters outputted by the reparametrization network for the $i$th molecule.
We rescale the gradients of the $i$th molecule with $\min \left(1, \frac{1}{s_i}\right)$.
This way small gradients are not scaled up but large gradients are scaled down.

\section{VMC optimization}\label{app:vmc}
A VMC step consists of three substeps: 1) sampling the square of the wave function $\psi_\theta^2$, 2) Computing the local energy $E_\theta(\evec)$ and gradients $\nabla_\theta E$ and 3) preconditioning the gradient with natural gradient descent $\mF^{-1}\nabla_\theta E$.
To sample the electronic wave function $\psi_\theta$, we use Metropolis-Hastings, i.e., in multiple iterations we perturb the electron positions from the last step with gaussian noise and perform rejection sampling based on the square of the wave function $\psi_\theta$.
Next, we compute the local energies for each electron configuration as in \Eqref{eq:energy} and use these samples to approximate the gradients with \Eqref{eq:gradient}.
Finally, we precondition the gradient with the inverse of the Fisher information matrix (FIM).
As the FIM scales quadratically with the number of parameters, realizing it and computing its inverse is infeasible.
Instead, we use the conjugate-gradient (CG) method to approximate its inverse~\citep{neuscammanOptimizingLargeParameter2012}.
For efficiency reasons, we compute the output of the reparametrization network once for the first two steps as it's constant throughout sampling and energy calculations.
Before applying the update, we clip the norm of the gradient to 1~\citep{pascanuDifficultyTrainingRecurrent2013} such that different system sizes do not require different choices of learning rates~\citep{gaoAbInitioPotentialEnergy2022}.

\section{Experimental setup}\label{app:setup}
\begin{table}
    \caption{Default hyperparameters.}\label{tab:hyperparameters}
    \centering
    \resizebox{\columnwidth}{!}{
        \begin{tabular}{llc}
            \toprule
            & Hyperparameter & Value \\
            
            \midrule
            Pretraining & Steps & 1e4 \\
            &Basis & STO-6G \\
            &Method & RHF \\
            
            \midrule
            Optimization & Steps & 6e4 \\
            &Learning rate & $\frac{0.1}{1+\frac{t}{100}}$ \\
            &Batch size & 4096\\
            &Damping & 1e-4 $\sigma[E_L]$\\
            &Local energy clipping & 5\\
            &Max grad norm & 1 \\
            &CG max steps & 100 \\
            
            \midrule
            MCMC & Target pmove & 0.5 \\
            &\# Steps & 40 \\
            
            \midrule
            \ourwf{}&Hidden dim & 256 \\
            &E-E int dim & 32 \\
            &Layers & 4 \\
            &Activation & SiLU\\
            &Determinants & 16 \\
            &Jastrow layers & 3 \\
            &Filter hidden dims & [16, 8] \\
            
            \midrule
            Reparametrization & Embedding dim & 128 \\
            &MLP layers & 4 \\
            &Message dim & 64 \\
            &Layers & 3 \\
            &Activation & SiLU \\
            &Filter hidden dims & [64, 16] \\
            
            \bottomrule 
        \end{tabular}
    }
\end{table}
We implemented all experiments and methods in JAX~\citep{bradburyJAXComposableTransformations2018}.
As we cannot rely on fixed tensor shapes like in previous work where only the spatial arrangements varied within a batch, we implemented everything with masking operations.
We generally parallelize all operations where possible over all molecules within a batch.
Exceptions are determinant calculations and the computation of the local energy.
While one can parallelize the determinant operation if one pads smaller matrices, we found this parallelization to be slower than performing the determinant calculations sequentially.
For computing the local energy, one needs to compute the Laplacian, i.e., the trace of the Hessian, of the log wave function.
This is a computationally demanding task where higher memory efficiency can be achieved by serializing across molecules and electrons.

For pretraining, we use the LAMB optimizer~\citep{youLargeBatchOptimization2020} while for VMC we use gradient descent with a maximal gradient norm of 1.
During VMC, we apply the gradient clipping from \citet{vonglehnSelfAttentionAnsatzAbinitio2023}, i.e., we clip all deviations from the median larger than 5 times the mean absolute deviation before computing the mean of the local energies.

All experiments ran on 1 to 4 Nvidia A100 GPUs depending on the system size.
If not otherwise specified, we use the hyperparameters from Table~\ref{tab:hyperparameters}.

\section{Molecular structures}\label{app:molecules}
Here, we list for each of our experiments the molecular structures and reference calculations.

For testing training on similar structures, we use the hydrogen rectangle from \citet{pfauInitioSolutionManyelectron2020}.
For the six-element hydrogen chain, we use the pretraining geometries from \citet{scherbelaSolvingElectronicSchrodinger2022}, and for the ten-element hydrogen chain, we use the geometries from \citet{mottaSolutionManyElectronProblem2017}.

The extended hydrogen chains for the extensivity experiment are generated by having a $n$-element chain of hydrogen atoms with interatomic distances of \SI{1.8}{\bohr}.

For testing dissimilar structures, we use the same distances for nitrogen as in \citet{pfauInitioSolutionManyelectron2020}.
As reference energy, we use twice the atomic energy of nitrogen from \citet{chakravortyGroundstateCorrelationEnergies1993} plus the experimental dissociation energy from \citet{leroyAccurateAnalyticPotential2006}.
For the additional hydrogen chain, we reuse the geometries from \citet{mottaSolutionManyElectronProblem2017}.
For ethene, we use the evaluation structures from \citet{scherbelaSolvingElectronicSchrodinger2022}.

In our transferability experiment, we take the six-element and ten-element hydrogen chain as well as the methane, and ethene structures and energies from \citet{scherbelaSolvingElectronicSchrodinger2022}.
The cyclobutadiene structures are from \citet{lyakhMultireferenceNatureChemistry2012} with the final VMC energies of \citet{gaoSamplingfreeInferenceAbInitio2023} as reference.

For benzene, we reuse the same geometry from \citet{renGroundStateMolecules2022} as previous works.

\section{Timings}\label{app:time}
\begin{table}
    \centering
    \caption{Forward pass timings of FermiNet, \ourwf{}, and \our{}.}\label{tab:timings}
    \begin{tabular}{crrr}
        \toprule
        (\# nuclei / \#electrons) & FermiNet & Moon & Globe\\
        \midrule
        1 / 10 & \SI{1.8}{\micro\second} & \SI{1.6}{\micro\second} & \SI{1.1}{\milli\second} \\
        1 / 40 & \SI{10.1}{\micro\second} & \SI{8.2}{\micro\second} & \SI{1.2}{\milli\second} \\
        1 / 80 & \SI{32.0}{\micro\second} & \SI{25.5}{\micro\second} & \SI{1.4}{\milli\second} \\
        10 / 10 & \SI{2.2}{\micro\second} & \SI{4.5}{\micro\second} & \SI{2.3}{\milli\second} \\
        10 / 40 & \SI{12.1}{\micro\second} & \SI{14.7}{\micro\second} & \SI{2.5}{\milli\second} \\
        10 / 80 & \SI{36.7}{\micro\second} & \SI{39.0}{\micro\second} & \SI{2.6}{\milli\second} \\
        20 / 40 & \SI{15.2}{\micro\second} & \SI{21.7}{\micro\second} & \SI{3.1}{\milli\second} \\
        20 / 80 & \SI{44.3}{\micro\second} & \SI{55.6}{\micro\second} & \SI{3.1}{\milli\second} \\
        \bottomrule
    \end{tabular}
    \vspace{-0.15in}
\end{table}
Table~\ref{tab:timings} lists the timings for the forward pass of FermiNet, \ourwf{}, and \our{}.
For systems with few nuclei, we find \ourwf{} to perform faster than FermiNet while reaching higher accuracies.
Though, this advantage reverses with an increasing number of nuclei due to the focus on electron-nuclei interactions.

While Globe's forward pass is significantly slower than FermiNet's or \ourwf{}'s it must only be executed once per step, i.e., sampling and energy computations do not require the reparametrization network but just the wave function.

\section{Parameters}\label{app:parameters}
\begin{table}
    \caption{Parameter counts for FermiNet, PsiFormer, \ourwf{}, and \our{}.} \label{tab:parameters}
    \centering
    \begin{tabular}{cccc}
        \toprule
        FermiNet & PsiFormer & \ourwf{} & Globe \\
        \midrule
        0.7M & 1.6M & 1M & 13M \\
        \bottomrule
    \end{tabular}
    \vspace{-0.15in}
\end{table}
In Table~\ref{tab:parameters}, we list the number of parameters for FermiNet~\citep{pfauInitioSolutionManyelectron2020}, PsiFormer~\citep{vonglehnSelfAttentionAnsatzAbinitio2023}, \ourwf{}, and Globe.
With its 300k more parameters, we found \ourwf{} to outperform FermiNet significantly in various benchmarks.
Compared to PsiFormer, we find \ourwf{} to perform similarly with 600k fewer parameters.

\our{}'s large number of parameters is mostly due to the lage output space, e.g., 10M are concentrated in a dense layer to predict the 8192-dimensional output space for the 32 $\tilde{w}_i^{k\delta}\in \R^{256}$ orbital embeddings, see Equation~\ref{eq:orbitals}.

\section{\ourwf{} size ablation}\label{app:ablation}
\begin{figure}
    \centering
    \includegraphics[width=\linewidth]{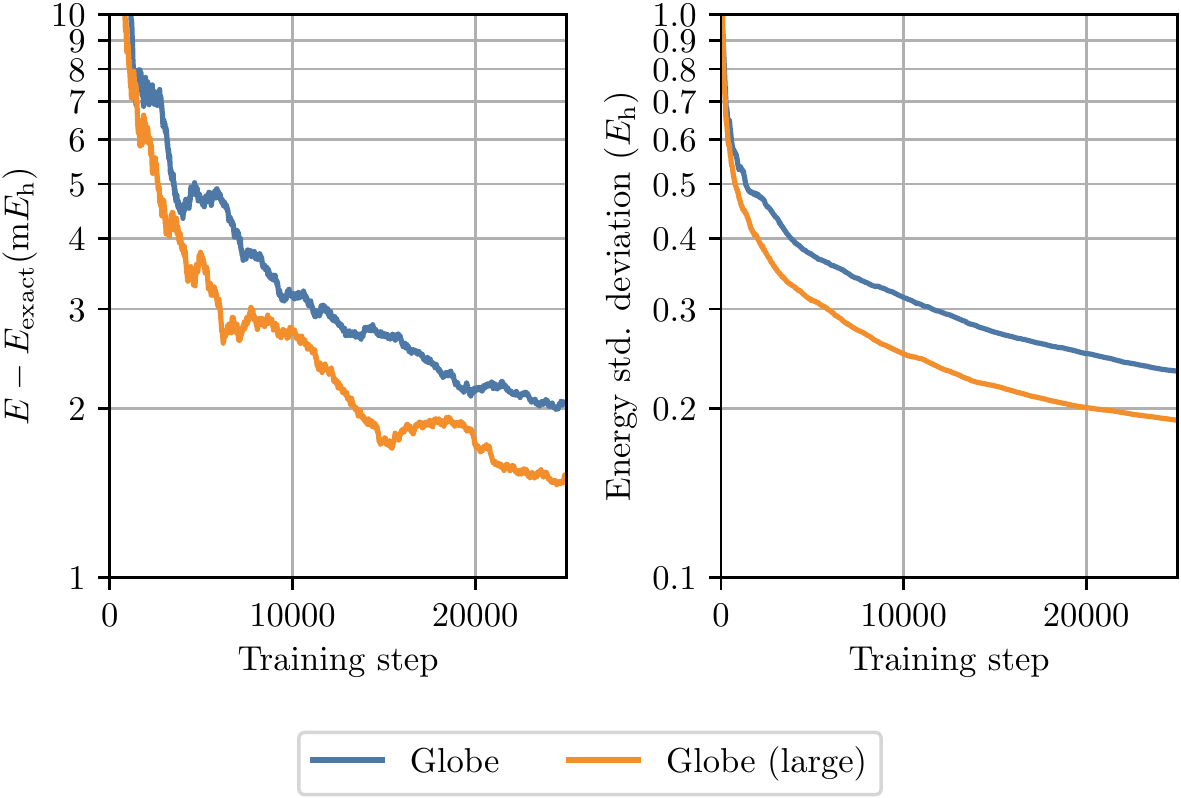}
    \vspace{-0.25in}
    \caption{Abalation of \ourwf{} where the hidden dimension has been increased to 512 and the number of determinants to 32. In agreement with previous work, we find that increasing the size of the wave function improves variational results accordingly~\citep{spencerBetterFasterFermionic2020,vonglehnSelfAttentionAnsatzAbinitio2023}.}\label{fig:ablation}
\end{figure}
While jointly training on diverse molecules seems to decrease the accuracy of neural wave functions, here we want to investigate the effect of the size of the network on training.
We perform the common augmentation of increasing the hidden dimension to 512 and the number of determinants to 32~\citep{spencerBetterFasterFermionic2020,vonglehnSelfAttentionAnsatzAbinitio2023} and compare the average energy on the diverse small molecule dataset from Section~\ref{sec:experiments}.

The energy during training in Figure~\ref{fig:ablation} agrees with previous results on neural network wave functions that increasing the network size increases accuracy~\citep{spencerBetterFasterFermionic2020,vonglehnSelfAttentionAnsatzAbinitio2023}.
As learning diverse molecular wave functions within a single neural network may require more parameters than learning a single wave function, we already see such improvements in small molecules.

\section{Standard deviation on hydrogen systems}\label{app:hydrogen_std}
\begin{figure}
    \includegraphics[width=\linewidth]{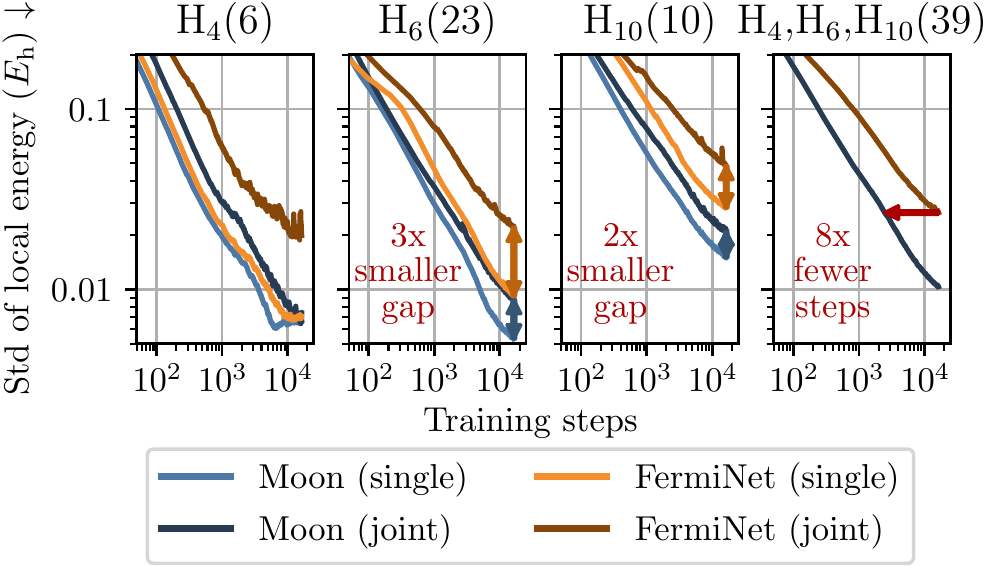}
    \vspace{-0.25in}
    \caption{
        Convergence plots of \our{} with \ourwf{} and FermiNet.
        Numbers in brackets show the number of geometries per molecule.
        In joint training, \ourwf{} converges 8 times faster and closes the gap to individual training by a factor of at least 2.
    }
    \label{fig:hydrogen_std}
    \vspace{-0.15in}
\end{figure}
While the energy of a system is a good indicator of convergence, any ground-state wave function will have no standard deviation in its local energy.
Thus, we can take a look at the standard deviation of the local energy as a proxy for the convergence of a wave function.
In Figure~\ref{fig:hydrogen_std}, we plot the standard deviation during the training on similar hydrogen systems.
We observe that \ourwf{} converges 8 times faster in joint training while also closing the gap to the individual trainings by a factor of at least 2.
\begin{figure*}
    \includegraphics[width=\linewidth]{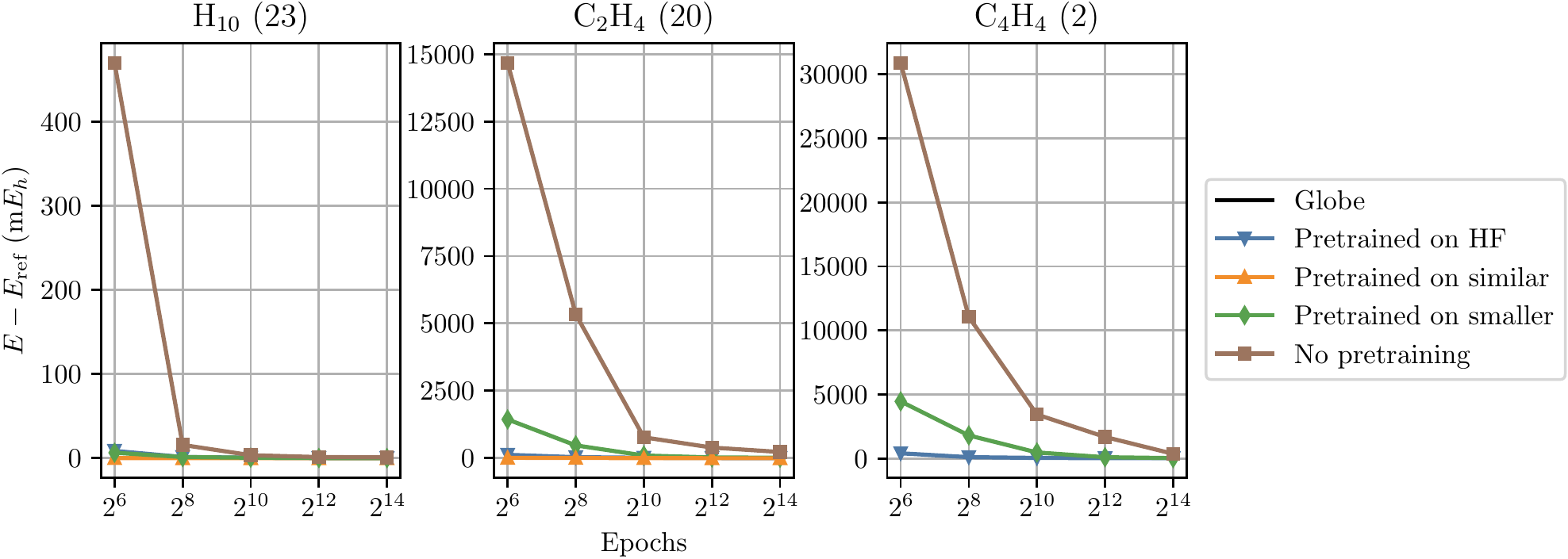}
    \vspace{-0.25in}
    \caption{
        Extended view of Figure~\ref{fig:transferibility} without a limit on the y-axis.
        Convergence of \our{} suffers 
    }
    \label{fig:transfer_ext}
    \vspace{-0.15in}
\end{figure*}

\begin{figure}
    \includegraphics[width=\linewidth]{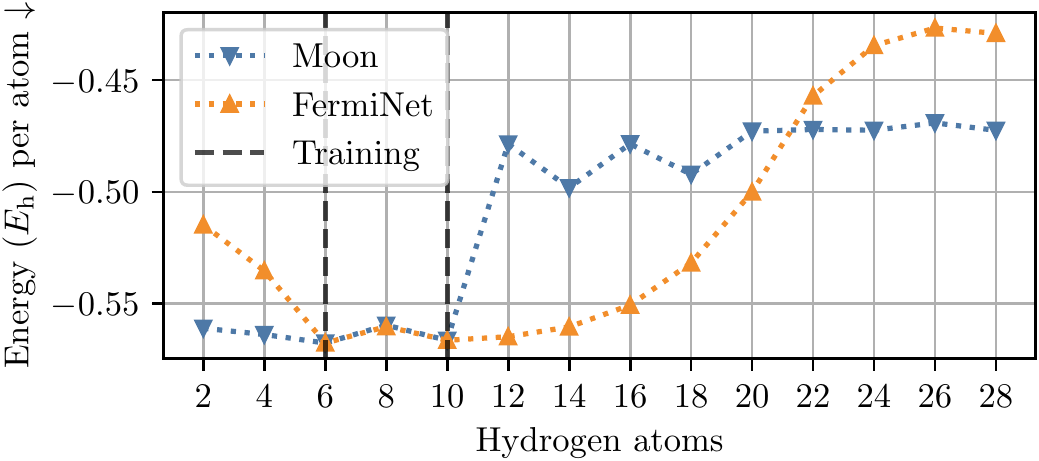}
    \vspace{-0.25in}
    \caption{
        Evaluation of \our{} with \ourwf{} and FermiNet, trained on small hydrogen clusters, on $n$-elment hydrogen chains.
        Thanks to \our{}'s localized orbitals, the energy per atom converges to a constant value for longer chains.
    }
    \label{fig:extensivity}
    \vspace{-0.15in}
\end{figure}
\section{Extensivity}\label{app:extensivity}
We analyze the behavior of \our{} with \ourwf{} and FermiNet on increasing lengthy hydrogen chains.
We first train \our{} on the same hydrogen structures as in the previous experiment, i.e., H$_4$, H$_{6}$, and H$_{10}$.
After training, we evaluate \our{} on smaller and larger $n$-element hydrogen chains.

Figure~\ref{fig:extensivity} depicts the scaling behavior of \ourwf{} and FermiNet depending on the system size.
As both are upper bounds to the true energy, lower energies are better.
Considering that no finetuning is done, neither FermiNet nor \ourwf{} diverges far from the trained energies per atom.
Interestingly, \ourwf{} performs better to smaller substructures like the hydrogen dimer H$_2$ but results in higher energies for moderately larger chains.
Thanks to its locality, \ourwf{}'s energy per atom is lower than FermiNet's with further increasing system sizes.

\section{Extended view on transferibility}\label{app:transfer}
As Figure~\ref{fig:transferibility} in Section~\ref{sec:experiments} does not include the training curves for ethene and cyclobutadiene without pretraining, we present an extended version in Figure~\ref{fig:transfer_ext}.
One can see that dropping pretraining impedes convergences.
Meanwhile, thanks to its graph-learned approach to molecular orbitals, \our{}, with pretraining on smaller molecules, is the first method to reach convergence on larger structures without performing a SCF calculation first.

\end{document}